\documentclass[11pt,x11names]{article}

\usepackage[T1]{fontenc}
\usepackage[latin1]{inputenc}

\usepackage{hyperref}

\usepackage{fancyhdr,cancel}

\usepackage{listings}
\usepackage[inline]{enumitem}
\usepackage{multicol,xcolor}

\usepackage{times}

\usepackage[hang,small,labelfont=bf,up,textfont=it,up]{caption}
\usepackage{subcaption}
\usepackage{booktabs}

\usepackage{amsmath,amsfonts,amsthm,amssymb,bm,dsfont}	
\setlist{nolistsep}

\usepackage{setspace}

\usepackage{graphicx}

\usepackage{adjustbox}

\usepackage[a4paper,top=3cm,bottom=3cm,left=2.25cm,right=2.25cm]{geometry}

\usepackage{centernot}

\usepackage{numprint}
\npthousandsep{,}\npthousandthpartsep{}\npdecimalsign{.}

\DeclareMathOperator*{\argmax}{argmax}
\renewcommand{\vec}[1]{\boldsymbol{#1}}
\newcommand{\mvec}[1]{\mathbf{#1}}
\newcommand{\abs}[1]{\vert{#1}\vert}
\newcommand{\norm}[1]{\|{#1}\|}

\newcommand{\Xd}{\hat{\mvec X}_{:d}}
\newcommand{\Xdd}{\hat{\mvec X}_{d:}}

\newcommand{\xid}{\hat{\vec x}_{i,:d}}
\newcommand{\xidd}{\hat{\vec x}_{i,d:}}
\newcommand{\der}[2]{\frac{\partial{#1}}{\partial{#2}}}

\providecommand{\keywords}[1]{{\small{\textbf{\textit{Keywords ---}} #1}}}

\newcommand{\titledoc}{Spectral clustering on spherical coordinates under the degree-corrected stochastic blockmodel}
\newcommand{\titleshort}{Spectral clustering on spherical coordinates under the degree-corrected stochastic blockmodel}

\usepackage{csquotes} 
\usepackage{natbib}

\usepackage{sectsty}
\partfont{\fontsize{17}{17}\selectfont}

\usepackage{thmtools}
\newtheorem{definition}{Definition}

\usepackage{authblk}

\usepackage[ruled,linesnumbered]{algorithm2e}
\DeclareMathOperator{\argmin}{argmin}

\hypersetup{colorlinks,linkcolor={blue},citecolor={blue},urlcolor={blue}}  
\numberwithin{equation}{section}

\usepackage{mathtools}
\mathtoolsset{showonlyrefs}

\author[1]{Francesco Sanna Passino}
\author[1]{Nicholas A. Heard}
\author[2]{Patrick Rubin-Delanchy}

\affil[1]{Department of Mathematics, Imperial College London}
\affil[2]{School of Mathematics, University of Bristol}

\date{}

\title{\huge\textbf{\titledoc}}
	
\begin{document}


\maketitle


\begin{abstract}
Spectral clustering is a popular method for community detection in network graphs: 
starting from a matrix representation of the graph,
the nodes are clustered on a low dimensional projection obtained from a truncated spectral decomposition of the matrix. Estimating correctly the number of communities and the dimension of the reduced latent space is critical for good performance of spectral clustering algorithms. 
Furthermore, many real-world graphs, such as enterprise computer networks studied in cyber-security applications, often display heterogeneous within-community degree distributions. 
Such heterogeneous degree distributions are usually not well captured by standard spectral clustering algorithms. 
In this article, a novel 
spectral clustering algorithm is proposed for community detection under the degree-corrected stochastic blockmodel. The proposed method is based on a transformation of the spectral embedding to spherical coordinates, and a novel modelling assumption in the transformed space. The method allows for simultaneous and automated selection of the number of communities and the latent dimension for 
spectral embeddings of graphs with uneven node degrees.
Results show improved performance over competing methods in representing  
computer networks. 
\end{abstract}


\section{Introduction} \label{intro_section}

Network data are commonly observed in a variety of scientific fields, representing, for example, interactions between neurons in the brain in biology, or connections between computers in communication technologies.
A fundamental problem in the statistical analysis of networks is the task of finding groups of similar nodes, known as \textit{community detection}.
Spectral clustering methods \citep{Ng01,vonLuxburg07} provide one of the most popular approaches for the community detection task.
Such techniques essentially consist of two steps: 
\begin{enumerate*}[label=(\roman*)]
\item spectrally embedding the graph adjacency matrix, or some transformation thereof, into a low dimensional space, and
\item apply a clustering algorithm, usually Gaussian mixture modelling (GMM) or $k$-means, in the low dimensional space. 
\end{enumerate*}

Spectral clustering algorithms can be used to obtain estimates of the community structure under a variety of classical network models.
The traditional model for community detection is the \textit{stochastic blockmodel} \citep[SBM,][]{Holland83}: each node in the network is assigned to one of $K$ communities, and the probability of a connection between two nodes only depends on their community memberships. Asymptotic theory suggests that 
embeddings arising from SBMs can be modelled using Gaussian mixture models (GMMs) \citep{RubinDelanchy17}.
This article mainly concerns the degree-corrected stochastic blockmodel \citep[DCSBM,][]{Karrer11}, which extends the SBM, allowing for heterogeneity in the within-community degree-distribution.
In DCSBMs, the probability of a connection 
depends on the community memberships,  
but is 
adjusted by node-specific 
degree-correction parameters. However, unlike the SBM, spectral embeddings under the DCSBM do not adhere to a GMM, since the communities are represented by \textit{rays}. 

In principle, DCSBMs appear to be particularly
suitable for modelling graphs arising from cyber-security applications, and in particular computer network flow data representing summaries of connections between Internet Protocol (IP) addresses, since machines within the same organisation  
tend to have different levels of activity depending on their purpose. 
Furthermore, in computer networks, the need for degree-correction seems most obvious when nodes are observed for different amounts of time, so that their connection probabilities 
scale with their \textit{``total time on test''}. For example, if a new node enters the network, it would be beneficial to identify its community, despite having very few connections.

The suitability of DCSBMs for community detection in cyber-security, our application of interest,
is demonstrated in Figure~\ref{sim_sbms}, where the within-community out-degree distributions arising from two simulated bipartite SBM (Figure~\ref{sbm_degree}) and DCSBM (Figure~\ref{dcsbm_degree}), 
are compared to the out-degree distribution of a real computer network 
(Figure~\ref{icl_degree}). A detailed description of the simulation is given in Section~\ref{results_icl}. 
The shape of the degree-distribution of the computer network resembles the simulated DCSBM much more closely than the SBM, suggesting that a degree-correction is required for correctly estimating the communities.

\begingroup

\begin{figure}[!t]
\centering
\begin{subfigure}[t]{.32\textwidth}
\centering
\caption{SBM}
\includegraphics[width=.95\textwidth]{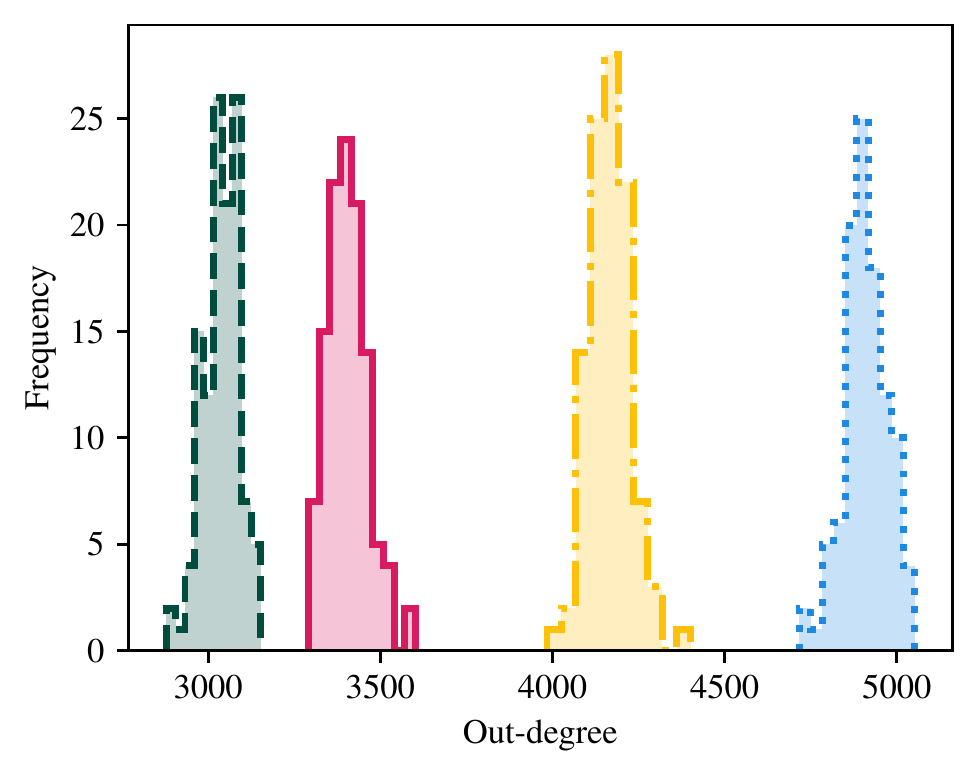}
\label{sbm_degree}
\end{subfigure}
\begin{subfigure}[t]{.32\textwidth}
\centering
\caption{DCSBM}
\includegraphics[width=.95\textwidth]{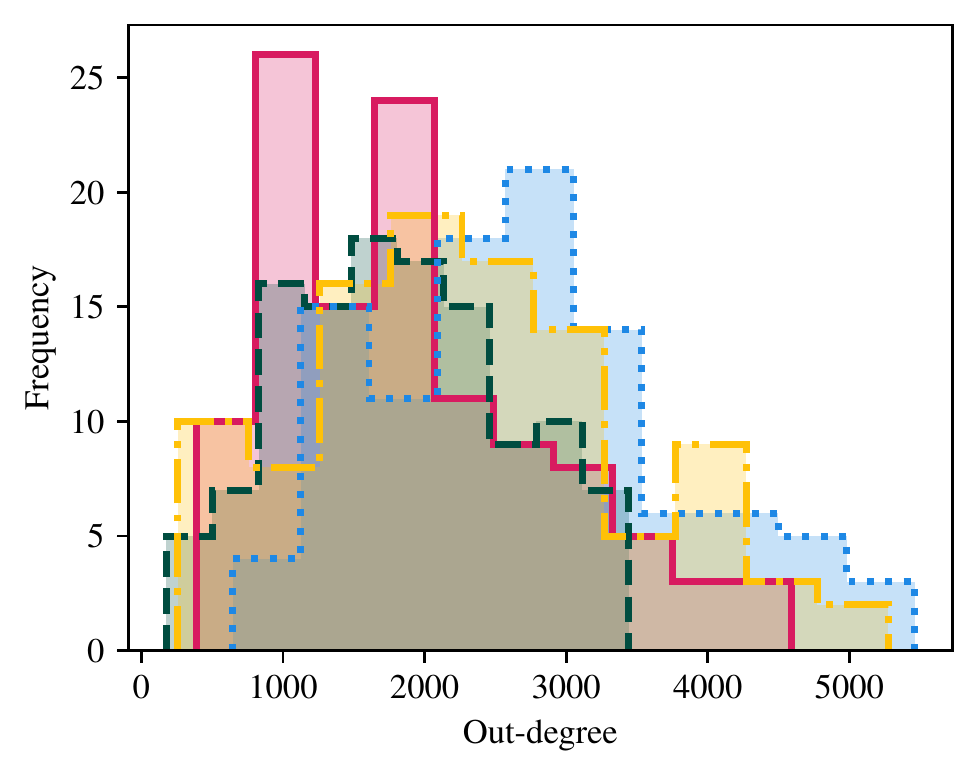}
\label{dcsbm_degree}
\end{subfigure}
\begin{subfigure}[t]{.32\textwidth}
\centering
\caption{ICL2, out-degree}
\includegraphics[width=.95\textwidth]{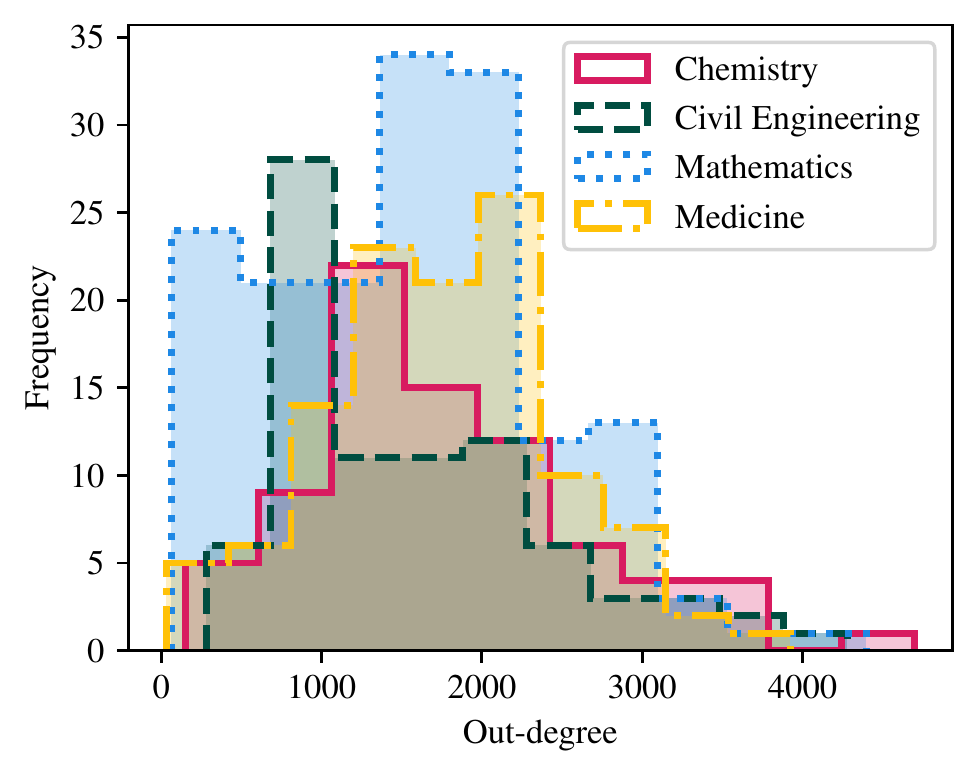}
\label{icl_degree}
\end{subfigure}
\caption{Histogram of within-community degree distributions from three bipartite networks with size $439\times\numprint{60635}$, obtained from \textit{(a)} a simulation of a SBM, \textit{(b)} a simulation of a DCSBM, and \textit{(c)} a real-world computer network (ICL2, cf. Section~\ref{results_icl}). 
} 
\label{sim_sbms}
\end{figure}
\endgroup

This article makes two main contributions. First, a novel spectral clustering 
algorithm for community detection under the DCSBM 
is proposed, based on a transformation of the 
embedding. 
Second, the proposed 
methodology is incorporated within a model selection framework for $d$, the embedding dimension, and $K$, the number of communities, providing a joint estimation method for those two parameters and the communities. 
The proposed method is shown to be competitive on simulated and real-world computer network data.

The article is structured as follows: Section~\ref{sec:background} describes the research question and related literature, followed by a preliminary discussion of spectral embedding techniques. Our proposed model is presented in Section~\ref{sec:model}. Parameter estimation and model selection are then discussed in Section~\ref{model_selection}. The proposed model is validated in Section~\ref{model_vali}, and results on simulated and real world computer network data are presented in Section~\ref{results_section}.

\section{Background and motivation} \label{sec:background}

A network can be  
expressed as a graph $\mathbb G=(V,E)$, consisting of a set of nodes $V$ of cardinality $n$, 
and a set of edges $E\subseteq V\times V$ representing the pairs of nodes which have interacted.
The graph is 
summarised by its adjacency matrix $\mvec A\in\{0,1\}^{n\times n}$, where $A_{ij}=\mathds 1_E\{(i,j)\}$ for $1\leq i,j\leq n$,
with 
$A_{ii}=0,\ 1\leq i\leq n$. If $(i,j)\in E \iff (j,i) \in E$, the graph is \textit{undirected}, implying that $\mvec A$ is symmetric; otherwise the graph is \textit{directed}.

The \textit{degree-corrected stochastic blockmodel} \citep[DCSBM,][]{Karrer11}  
is a popular model for community detection in graphs. 
For $K$ communities, the nodes are divided into blocks by random assignment of community membership indicators $\vec z=(z_1,\dots,z_n)\in\{1,\dots,K\}^n$, with community probabilities $\bm\psi=(\psi_1,\dots,\psi_K)$, $\sum_{j=1}^K\psi_j=1$. 
Furthermore, each node is assigned a degree-correction parameter $\rho_i\in[0,1]$. 
Each entry of the adjacency matrix is then independently modelled as 

\begingroup
\begin{equation}
  A_{ij} \sim \mathrm{Bernoulli}(\rho_i\rho_j B_{z_iz_j}),
  \label{dcsbm}
\end{equation}
\endgroup
where $\mvec B\in[0,1]^{K\times K}$ is a $K\times K$ matrix of probabilities such that $B_{k\ell}$ is 
a baseline probability for a node from community $k$ interacting with a node from community $\ell$.

Beyond 
DCSBMs, random dot product graphs \citep[RDPG,][]{Young07} represent a wider and more flexible class of models for network data. Each node is assigned a $d$-dimensional latent position $\vec x_i\in\mathbb R^d$ such that $\vec x_i^\intercal\vec x_j\in[0,1]$ for $i,j\in\{1,\dots,n\}$. The probability of a link between $i$ and $j$ is then determined as
$
A_{ij}\sim\mathrm{Bernoulli}(\vec x_i^\intercal\vec x_j).
$ 
The latent positions can be arranged in a matrix $\mvec X=[\vec x_1,\dots,\vec x_n]^\intercal\in\mathbb R^{n\times d}$ such that $\mathbb E(\mvec A)=\mvec X\mvec X^\intercal$.
For a positive definite block connectivity probability matrix $\mvec B$, 
DCSBMs can be expressed as RDPGs.  
If each community is assigned an \textit{uncorrected} position ${\bm\mu}_k\in\mathbb{R}^d$, such that $B_{k\ell}=\bm\mu_k^\intercal\bm\mu_\ell,\ k,\ell\in\{1,\dots,K\}$, the 
DCSBM is obtained by setting $\vec x_i=\rho_i\bm\mu_{z_i},\ i\in V$, conditional on the communities $\vec z$ and degree-correction parameters $\bm\rho=(\rho_1,\dots,\rho_n)$.

This article is primarily concerned with a novel technique for estimating the underlying node communities given an adjacency matrix,  
under a RDPG  
interpretation of the DCSBM. A joint estimation method is proposed for the community structure $\vec z$, the number of communities $K$, and the latent dimension $d$ of the latent positions. 

\subsection{Related literature, shortcomings, and proposed solutions}

Community detection based on  
DCSBMs is an active field of research. \cite{Zhao12} present a  
theory for assessing consistency under the DCSBM. \cite{Amini13} use a pseudo-likelihood approach, providing consistency results for the estimators. \cite{Peng16} frame the DCSBM in a Bayesian setting, using a logistic regression formulation with node correction terms.  \cite{Chen18} propose a convexified modularity maximisation approach. \cite{Gao18} obtain a minimax risk result for community detection in DCSBMs and propose a two-step clustering algorithm based on $k$-medians. 

Spectral clustering methods have emerged as one of the most popular approaches for community detection under the DCSBM \citep{Lei15,Gulikers17}.
A common technique 
uses $k$-means on the normalised rows of the embedding \citep{Qin13} obtained from the spectral decomposition of the regularised Laplacian matrix \citep{Chaudhuri12}. 
The row-normalisation of the embedding is a well-established approach for spectral clustering \citep{Ng01} under the DCSBM, 
but the normalised rows live in a $d-1$ dimensional manifold, so it  
is not fully appropriate to fit a model for $d$-dimensional clusters to such an embedding.
Alternative methods include the SCORE algorithm of \cite{Jin15}, which proposes to use $k$-means on an embedding scaled by the leading eigenvector, showing that the effect of degree heterogeneity can be largely removed.  

This article proposes a novel methodology for spectral clustering under the DCSBM, interpreted as a special case of RDPG:  
the $d$-dimensional spectral embedding is reduced to a set of $d-1$ \textit{directions}, or \textit{angles}, changing from a Cartesian coordinate system to a spherical system. 
This choice of transformation is carefully motivated by asymptotic theoretical properties of the embeddings arising from DCSBMs.

Additionally, many estimation methods commonly require the number of communities $K$ to be known.
Furthermore, spectral clustering methods require the specification of the embedding dimension $d$, and clustering is usually carried out \textit{after} selecting this parameter. This sequential approach is suboptimal, since the clustering configuration and the number of communities $K$ would ideally be estimated jointly with $d$. 
In practice, selecting $d$ and $K$ is a difficult task. 
\cite{SannaPassino19} and \cite{Yang19} independently proposed an automatic model selection framework for both the number of communities $K$ and the dimension $d$ of the latent node positions in SBMs, interpreted as RDPGs. 
In this work, the methodology is extended to DCSBMs, providing an
algorithm for practitioners.

\subsection{Spectral embedding} \label{defi_section}

Given a network adjacency matrix, spectral embedding methods provide estimates $\hat{\mvec X}$ of the latent positions $\mvec X$ in RDPGs, from decompositions 
of the adjacency matrix or its 
Laplacian. This article will mainly discuss the \textit{adjacency spectral embedding}, defined below.

\begin{definition}[Adjacency spectral embedding]
Consider a symmetric adjacency matrix $\mvec A\in\{0,1\}^{n\times n}$ and a positive integer $d\in\{1,\dots,n\}$. The adjacency spectral embedding (ASE) of $\mvec A$ in $\mathbb R^d$ is $\hat{\mvec X}=\hat{\bm\Gamma}\abs{\hat{\bm\Lambda}}^{1/2}$, where $\abs{\hat{\bm\Lambda}}$ is a diagonal $d\times d$ matrix containing on the main diagonal the absolute value of the top-$d$ eigenvalues of $\mvec A$ in magnitude, in decreasing order, and $\hat{\bm\Gamma}$ is a $n\times d$ matrix containing corresponding orthonormal eigenvectors. 
\end{definition}

For a directed graph, the RDPG model assumes each node $i\in V$ has two latent positions $\vec x_i,\vec x_i^\prime\in\mathbb R^d$,  such that $A_{ij}\sim\mathrm{Bernoulli}(\vec x_i^\intercal\vec x_j^\prime)$.
The corresponding spectral embedding uses singular value decomposition. 
Bipartite graphs can be interpreted as a special case of directed graphs, and therefore are also spectrally embedded using the same procedure. 

\begin{definition}[Directed adjacency spectral embedding] \label{dase}
Consider an adjacency matrix $\mvec A\in\{0,1\}^{n\times n}$, not necessarily symmetric, and a positive integer $d\in\{1,\dots,n\}$. The directed adjacency spectral embedding (DASE) of $\mvec A$ in $\mathbb R^d$ is jointly given by $\hat{\mvec X}=\hat{\mvec U}\hat{\mvec S}^{1/2}$ and $\hat{\mvec X}^\prime=\hat{\mvec V}\hat{\mvec S}^{1/2}$, where $\hat{\mvec S}$ is a diagonal $d\times d$ matrix containing on the main diagonal the top-$d$ singular values of $\mvec A$ in magnitude, in decreasing order, and $\hat{\mvec U}$ and $\hat{\mvec V}$ are $n\times d$ matrices containing corresponding orthonormal left and right singular vectors respectively. 
\end{definition}
 
\subsection{Asymptotic properties of spectral embedding of DCSBMs}

An important result in the RDPG literature establishes the rows of the ASE as consistent estimators of the latent positions 
\citep{Sussman14}. Furthermore, central limit theorems \citep[ASE-CLTs,][]{Athreya16,RubinDelanchy17,Tang18} provide strong justification for estimation of 
the latent positions via ASE. 

When ASE is applied to DCSBMs, the asymptotic theory \citep[see, for example,][]{RubinDelanchy17} predicts that 
each community 
is represented as a \textit{ray} from the origin in the embedding space.
An example is given in Figure~\ref{example_dcsbm}, which shows the two-dimensional ASE for a simulated DCSBM with $n=\numprint{1000}$ nodes and $K=4$ communities.

\begingroup

\begin{figure}[!t]
\centering
\begin{subfigure}[t]{.495\textwidth}
\centering
\caption{$K=4$, coloured by community membership}
\includegraphics[width=.975\textwidth]{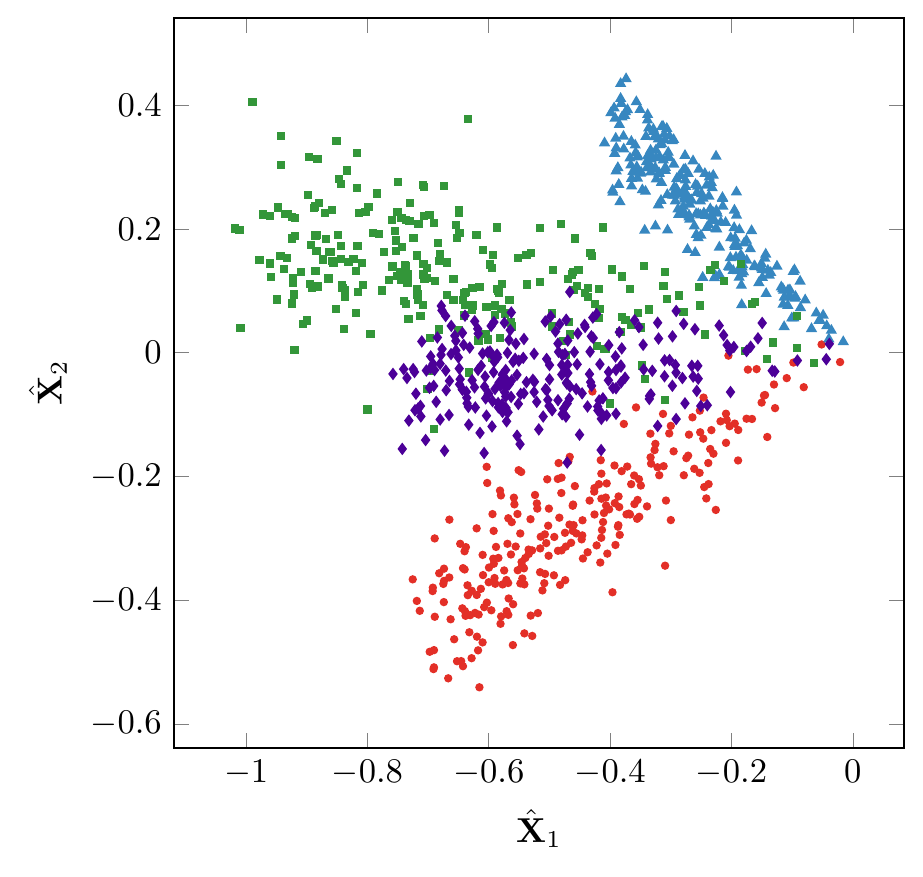}
\label{example_dcsbm}
\end{subfigure}
\begin{subfigure}[t]{.495\textwidth}
\centering
\caption{$K=2$, with ASE-CLT contours}
\includegraphics[width=.93\textwidth]{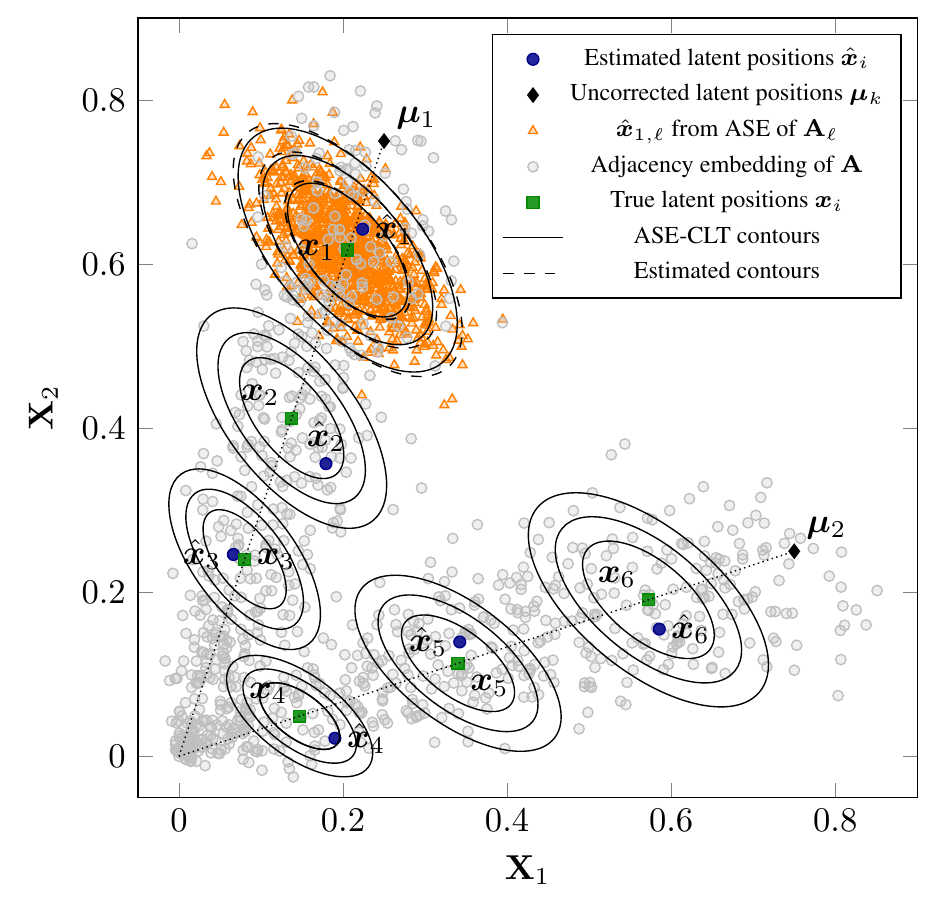}
\label{ellipses}
\end{subfigure}
\caption{Scatterplots of the two-dimensional ASE of a simulated DCSBM with (a) $K=4$, and (b) $K=2$. Figure~\ref{ellipses} also highlights the true and estimated latent position for 6 nodes, with the corresponding 50\%, 75\% and 90\% contours from the ASE-CLT, and the estimated latent positions $\hat{\vec x}_{1,\ell}$ for $\vec x_1$ from simulated DCSBM adjacency matrices $\mvec A_\ell,\ \ell=1,\dots,1000$.}
\label{example_embedding}
\end{figure}
\endgroup
 
To describe each community, let $\vec x=\rho\bm\mu_k\in\mathbb R^d$ be the underlying latent position for a node in community $k$. Further, let $\hat{\vec x}^{(n)}$ be the ASE estimator of $\vec x$, obtained from a graph with $n$ nodes.
For $d$ fixed and known, the ASE-CLT \citep[for example,][]{Athreya16}, applied to DCSBMs, establishes that 
\begin{equation}
\lim_{n\to\infty} \mathbb P\left\{\sqrt{n}\left(\mvec Q^{(n)}\hat{\vec x}^{(n)} - \vec x\right) \leq \vec v\mid \vec x=\rho\bm\mu_k \right\}\to\Phi_d\{\vec v,\bm\Sigma_k(\rho)\}, \label{ase-clt}
\end{equation}
where $\vec v\in\mathbb R^d$, $\mvec Q^{(n)}\in\mathbb R^{d\times d}$ is an orthogonal matrix, $\bm\Sigma_k(\rho)$ is a $d\times d$ community-specific covariance matrix depending on $\rho$, and $\Phi_d\{\cdot,\bm\Sigma\}$ is the CDF of a $d$-dimensional Gaussian distribution with zero mean and covariance $\bm\Sigma$.  \cite{Jones20} extended the result \eqref{ase-clt} to the DASE (\textit{cf.} Definition~\ref{dase}). 
In simpler terms, \eqref{ase-clt} implies that, for $n$ large, 
the estimated latent position $\hat{\vec x}_i$ is normally distributed about $\rho_i\bm\mu_{z_i}$, after a suitable orthogonal transformation has been applied to the embedding (accounting both for the ambiguity in the choice of eigenvectors or singular vectors within the spectral embedding procedure, and the latent position identifiability in the RDPG).
The theorem is exemplified by Figure~\ref{ellipses}, which displays the two-dimensional ASE of a simulated DCSBM with $n=\numprint{1000}$ and $K=2$ equally probable communities, with $\bm\mu_1=[1/4,3/4]$, $\bm\mu_2=[3/4,1/4]$, and $\rho_i\sim\mathrm{Uniform}(0,1)$. For 6 nodes, the plot highlights the true and estimated latent positions, and corresponding theoretical Gaussian contours from the ASE-CLT. 
Additionally, the simulation of the adjacency matrix is repeated $\numprint{1000}$ times using the \textit{same} true underlying latent positions as the first simulation. Then, the ASE-estimated latent position for the first node is plotted for each simulation, along with the Gaussian contours estimated from the $\numprint{1000}$ estimates. The empirical contours (dashed lines) remarkably correspond to the theoretical ASE-CLT contours (solid lines) around the true latent position $\vec x_1$. Also, Figure~\ref{ellipses} shows that, within the same community, the \textit{true} latent positions all have the \textit{same spherical coordinates}, or \textit{angle to the origin}, whereas their corresponding \textit{ASE estimates} are distributed \textit{around} the line of the true latent positions, forming a \textit{community-specific ray}. 
The ASE-CLT \eqref{ase-clt} also establishes that estimated latent positions tend to asymptotically concentrate increasingly tightly around the rays connecting the origin and the unnormalised latent positions $\bm\mu_k,\ k\in\{1,\dots,K\}$.
Therefore, Figure~\ref{example_embedding} intuitively motivates the novel modelling choice proposed in this paper: estimating the node communities from the \textit{spherical coordinates}, or \textit{angles}, obtained from the ASE.  
The use of alternative coordinate systems for network analysis has been previously shown to have beneficial properties \citep[see, for example,][]{Krioukov10,Braun11,McCormick15,Lobato16}. Furthermore, a central limit theorem for the spherical coordinates of the latent positions is proved for $d=2$ in Appendix~\ref{asymptotic}, further establishing the suitable properties of such a transformation of the embedding.

 One of the main characteristics of the proposed methodology will be to allow for an initial misspecification of the parameter $d$, choosing an $m$-dimensional embedding with $m\geq d$, and then recovering the correct latent dimension by proposing a discriminative model for the extended embedding. In the remainder of the article, 
the notation $\Xd$ denotes 
the first $d$ columns of $\hat{\mvec X}$, and $\Xdd$ denotes the $m-d$ remaining columns.
Similarly, $\xid$ represents the first $d$ components $(\hat x_1,\dots,\hat x_d)$ of the vector $\hat{\vec x}_i$, and $\xidd$ the last $m-d$ components $(\hat x_{d+1},\dots,\hat x_{m})$.
Also, the \textit{row-normalised embedding} 
is denoted as $\tilde{\mvec X}=[\tilde{\vec x}_1,\dots,\tilde{\vec x}_n]^\intercal$, where $\tilde{\vec x}_i=\hat{\vec x}_i/\norm{\hat{\vec x}_i}$.
Importantly, the parameter $m$ 
is always assumed to be fixed.

\section{Modelling a transformation of DCSBM embeddings} \label{sec:model}

Consider an $m$-dimensional vector $\vec x\in\mathbb R^m$. The $m$ Cartesian coordinates $\vec x=(x_1,\dots,x_m)$ can be converted in $m-1$ spherical coordinates ${\bm\theta}=(\theta_1,\dots,\theta_{m-1})$ on the unit $m$-sphere 
using a mapping $f_m:\mathbb R^m\to[0,2\pi)^{m-1}$ such that $f_m:\vec x\mapsto{\bm\theta}$, where:
\begin{gather}
\theta_1 = \left\{\begin{array}{ll} \arccos(x_2/\|\vec x_{:2}\|) & x_1\geq 0, \\ 2\pi-\arccos(x_2/\|\vec x_{:2}\|) & x_1< 0, \end{array} \right. \label{cart_to_sphere1}\\
 \theta_j = 2\arccos(x_{j+1}/\|\vec x_{:j+1}\|),\ j=2,\dots,m-1, 
\label{cart_to_sphere}
\end{gather}
where $\|\cdot\|$ is the Euclidean norm.

Consider an $(m+1)$-dimensional adjacency embedding $\hat{\mvec X}\in\mathbb R^{n\times(m+1)}$ and define its transformation $\hat{\bm\Theta}=[\hat{\bm\theta}_1,\dots,\hat{\bm\theta}_n]^\intercal\in[0,2\pi)^{n\times m}$, where $\hat{\bm\theta}_i=f_{m+1}(\hat{\vec x}_i),\ i=1,\dots,n$. The symbols $\hat{\bm\Theta}_{:d}$ and $\hat{\bm\theta}_{i,:d}$ will denote respectively the first $d$ columns of the matrix and $d$ elements of the vector, and $\hat{\bm\Theta}_{d:}$ and $\hat{\bm\theta}_{i,d:}$ will represent the remaining $m-d$ components.   

In this article, a model is proposed for the transformed embeddings $\hat{\bm\Theta}$. 
Suppose a latent space dimension $d$, $K$ communities, and latent community assignments $\vec z=(z_1,\dots,z_n)$. 
The transformed coordinates $\hat{\bm\Theta}$ are assumed to be generated independently from community-specific $m$-dimensional multivariate normal distributions: 

\begingroup
\begin{equation}
\hat{\bm\theta}_i \vert d,z_i,\bm\vartheta_{z_i},\bm\Sigma_{z_i}, \bm\sigma^2_{z_i} \sim \mathbb N_m \left( \begin{bmatrix} \bm\vartheta_{z_i} \\ \pi\mvec 1_{m-d} \end{bmatrix}, \begin{bmatrix} \mvec\Sigma_{z_i} & \vec 0 \\ \vec 0 & \vec\sigma^2_{z_i}\mvec I_{m-d} \end{bmatrix} \right). \label{full_model}
\end{equation}
\endgroup
where $\bm\vartheta_k\in{[0,2\pi)}^d$, $k=1,\dots,K$, represents a community-specific mean angle, $\mvec 1_m$ is a $m$-dimensional vector of ones, $\bm\Sigma_k$ is a $d\times d$ full covariance matrix, and $\bm\sigma^2_k=(\sigma^2_{k,d+1},\dots,\sigma^2_{k,m})$ is a vector of positive variances. 
The model in \eqref{full_model} could be also completed in a Bayesian framework using the same prior distributions chosen in \cite{SannaPassino19}. 
For fixed $d$ and $K$, consider a set of mixing proportions $\bm\psi=(\psi_1,\dots,\psi_K)$ such that $\mathbb P(z_i=k)=\psi_k$, where $\psi_k\geq0, k=1,\dots,K,$ and $\sum_{k=1}^K\psi_k=1$. 
After marginalising out $\vec z$, the likelihood function is:
\begin{equation}
L(\hat{\bm\Theta}\vert d,K) = \prod_{i=1}^n \left\{ \sum_{j=1}^K \psi_j \phi_d(\hat{\bm\theta}_{i,:d};\bm\vartheta_j,\bm\Sigma_j) \phi_{m-d}(\hat{\bm\theta}_{i,d:};\pi\vec 1_{m-d},\bm\sigma^2_j\mvec I_{m-d}) \right\},
\label{incomplete_likelihood}
\end{equation}
where $\phi_q(\cdot;\bm\vartheta,\bm\Sigma)$ is the density of a $q$-dimensional normal distribution with mean $\bm\vartheta$ and variance $\bm\Sigma$.
Note that, normally, a wrapped normal distribution would be preferred for circular data. 
On the other hand, in the context of DCSBMs, it is known that $\hat{\bm\Theta}$ arises from a transformation of the embedding $\hat{\mvec X}$, and the form of the transformation can be used to inform the modelling decisions. The arccosine function is monotonically decreasing, and communities will tend to have similar values 
in $\hat x_{i,j}/\|\hat{\vec x}_{i,:j}\|$ for $j=1,\dots,m-1$, see \eqref{cart_to_sphere}.
If two points in $\hat{\bm\Theta}$ reach the extremes $0$ and $2\pi$, 
then they are unlikely to belong to the same community. Therefore,  
wrapped distributions do not apply to this context. 

The only case that could cause concern is $\theta_{i,1}$, where the transformation \eqref{cart_to_sphere1} is not monotonic, but has a discontinuity at $0$. 
The first column $\hat{\mvec X}_1$ 
of the ASE 
corresponds to the scaled leading eigenvector of the adjacency matrix, and therefore its elements have \textit{all} the same sign by the Perron-Frobenius theorem for non-negative matrices \citep[see, for example,][]{Meyer00}. The theorem 
makes the transformation 
$\vec x_{:2}\mapsto\theta_1$ in \eqref{cart_to_sphere1} monotonic, since one of the two conditions in the equation for $\theta_1$ is satisfied by all the values in $\hat{\mvec X}_1$. 

The rationale behind the model assumptions in \eqref{full_model} is to utilise the method of normalisation to the unit circle \citep{Qin13} but assume normality on the spherical coordinates, not on their Cartesian counterparts, as discussed in the comments to Figure~\ref{ellipses}.
The transformed initial components $\hat{\bm\theta}_{i,:d}$ are assumed to have unconstrained mean vector $\bm\vartheta_k\in{[0,2\pi)}^d$, and a positive definite community-specific $d\times d$ covariance matrix $\mvec\Sigma_k$.  

In contrast to the structured model for $\hat{\bm\Theta}_{:d}$, the remaining $m-d$ dimension of the embedding are modelled as noise, using similar constraints on $\hat{\vec x}_{i,d:}$ to those imposed in \cite{SannaPassino19} and \cite{Yang19}: the mean of the distribution is a $(m-d)$-dimensional vector centred at $2\arccos(0)=\pi$, and the covariance is a diagonal matrix $\vec\sigma^2_k\mvec I_{m-d}$ with positive diagonal entries. The mean value of $\pi$ reflects the assumption in \cite{SannaPassino19} and \cite{Yang19} of clusters centred at zero: for $j>d$, $x_{i,j}$ is expected to be near $0$, 
which makes the transformed coordinate centre fluctuate around $\pi$. 
Importantly, the assumption of cluster-specific variances on $\hat{\bm\Theta}_{d:}$ implies that $d$ does \textit{not} have the simple interpretation of being the number of dimensions relevant for clustering. This fundamentally differentiates the proposed modelling framework from traditional variable selection methods within clustering \citep[see, for example,][]{Dean06}. 
The parameter $d+1$ in this model represents the dimension of the latent positions that generate the network, or 
the rank of the block connectivity matrix $\mvec B$.

It must be further remarked that the model \eqref{full_model} concerns the distribution of the spherical coordinates of the ASE estimator of the DCSBM latent positions, and it is \textit{not} a generative model for DCSBMs. If used as a generative model, paired with a distributional assumption on the degree-correction parameters, the model \eqref{full_model} would generate a \textit{noisy} DCSBM, where the underlying latent positions are scattered around the rays, and not perfectly \textit{on} the rays as a traditional DCSBM (\textit{cf.} Figure~\ref{ellipses}). Considering the ASE-CLT \eqref{ase-clt}, the estimation procedure proposed in this work would still be applicable also to such a noisy DCSBM.

Finally, note that the method relies on an initial choice of the embedding dimension $m\geq d$. The parameter should be large enough to avoid potential issues with the case $m<d$. Choosing $m$ is arguably easier than choosing $d$, and in principle one could pick $m=n$. As a rule of thumb, the parameter could be chosen as the third or fourth elbow based on the criterion of \cite{Zhu06}. Note that, for $m\geq d$, $\hat{\bm\Theta}_{:d}$ is invariant to the choice of $m$: embeddings $\hat{\bm\Theta}_{:d}$ calculated from $m$-dimensional and $m^\ast$-dimensional embeddings, with $m\neq m^\ast$, are identical. This is not the case for the row-normalised embedding: embeddings $\tilde{\mvec X}_{:d}$ obtained from $m$-dimensional and $m^\ast$-dimensional embeddings are in general \textit{not} equal. 

\section{Model selection and parameter estimation} \label{model_selection}

For selection of the number of communities $K$ and latent dimension $d$, a classical approach of model comparison via information criteria is adopted, 
already used by \cite{Yang19} in the context of simultaneous model selection in SBMs. 
Suppose that the maximum likelihood estimate (MLE) of the model parameters for fixed $d$ and $K$ is $\{\hat{\bm\vartheta}_k,\hat{\bm\Sigma}_k,\hat{\bm\sigma}^2_k,\hat\psi_k\}_{k=1,\dots,K}$. 
The Bayesian Information Criterion (BIC) is defined as:
\begin{multline}
\text{BIC}(d,K) = 
-2\sum_{i=1}^n \log\left\{\sum_{k=1}^K \hat\psi_j \phi(\hat{\bm\theta}_{i,:d};\hat{\bm\vartheta}_k,\hat{\bm\Sigma}_k) \prod_{j=d+1}^m \phi(\hat\theta_{i,j};\pi,\hat\sigma^2_{k,j}) \right\} \\
+ K\log(n)(d^2/2+d/2+m+1). 
\label{bic}
\end{multline} 
The first term of the BIC \eqref{bic} is the \textit{negative log-likelihood} \eqref{incomplete_likelihood}, whereas the second term is a penalty. 
The estimates of $d$ and $K$ will correspond to the pair that minimises \eqref{bic}, obtained using grid search: $(\hat d,\hat K)=\argmin_{(d,K)}\text{BIC}(d,K)$. The latent dimension $d$ has range $\{1,\dots,m\}$, whereas $K\in\{1,\dots,n\}$. In practice, it is convenient to fix a maximum number of clusters $K^\ast$ in the grid search procedure, such that $K\in\{1,\dots,K^\ast\}$.

From \eqref{bic}, it follows that the MLE 
of the Gaussian model 
parameters 
is required for each pair $(d,K)$. 
The expectation-maximisation \citep[EM,][]{Dempster77} algorithm is typically used for problems involving likelihood maximisation in model based clustering \citep[for example,][]{Raftery02}. Finding  
the MLE for \eqref{incomplete_likelihood} only requires a simple modification of the standard algorithm for GMMs, 
adding constraints on the means and covariances of the last $m-d$ components. 
Under the assumption that the model for the spherical coordinates is correctly specified, then the same framework described in Theorem 1 in \cite{Yang19} could be used to obtain theoretical guarantees on the estimates.

Given the maximum likelihood estimates for $(d,K)$, 
the 
communities are estimated as 
\begin{equation}
\hat z_i=\argmax_{j\in\{1,\dots,\hat K\}}\left\{ \hat\psi_j \phi_{\hat d}\left(\hat{\bm\theta}_{i,:\hat d};\hat{\bm\vartheta}_j,\hat{\bm\Sigma}_j\right) 
\right\},\ i=1,\dots,n, \label{cluster_estimate}
\end{equation}
where a $\hat K$-component GMM is fitted to the $\hat d$-dimensional embedding, and the second Gaussian term in the likelihood \eqref{incomplete_likelihood}, accounting for the last $m-d$ components of the embedding, is removed to reduce the bias-variance tradeoff \citep{Yang19}.

For a given pair $(d,K)$, fast estimation and convergence can be achieved by initialising the EM algorithm with the 
MLE of a GMM 
fitted on the initial $d$ components of the $m$-dimensional embedding. 
This approach for initialisation will be followed in Section~\ref{results_icl}. The full procedure is summarised in Algorithm~\ref{dcsbm_algo}.
It must be remarked that the grid search procedure for estimating $(d,K)$ from \eqref{bic} requires $m\times K^\ast$ EM-algorithms. This is computationally intensive, especially if the number of nodes is large, or $m$ and $K^\ast$ are large. Therefore, the methodology for estimation of $(d,K)$ is 
 not scalable to very large graphs. 

\begingroup

\begin{algorithm}[!h]
\SetAlgoLined
\KwIn{
adjacency matrix $\mvec A\in\{0,1\}^{n\times n}$, maximum number of communities $K^\ast$.}
\KwResult{estimated community allocations $\hat{\vec z}$ and estimated latent dimensionality $\hat d$.}
select $m$ as the
third or fourth elbow of the scree plot \citep{Zhu06}, \\
calculate the $m$-dimensional ASE $\hat{\mvec X}\in\mathbb R^{n\times m}$ (or DASE for directed graphs), \\
transform the embedding $\hat{\mvec X}\in\mathbb R^{n\times m}$ into spherical coordinates $\hat{\bm\Theta}\in[0,2\pi)^{n\times(m-1)}$, \\
 \For{$d=1,2,\dots,m$}{
 \For{$K=1,\dots,K^\ast$}{
  calculate MLE $\{\hat\psi_j,\hat{\bm\vartheta}_j,\hat{\bm\Sigma}_j,\hat{\bm\sigma}^2_j\}_{k=1,\dots,K}$ of model \eqref{full_model} using the EM algorithm, \\  
  calculate $\text{BIC}(d,K)$ using \eqref{bic}, \\
  }
 }
 obtain the estimate $(\hat d,\hat K)=\argmin_{(d,K)}\text{BIC}(d,K)$ for the pair $(d,K)$, \\
 fit a $\hat K$-component GMM to 
 $\hat{\bm\Theta}_{:\hat d}$, and 
 estimate the communities 
 $\vec z$ using \eqref{cluster_estimate}. 
 \caption{Estimation of the latent dimension $d$, number of communities $K$, and community allocations $\vec z$, given a graph adjacency matrix $\mvec A$.}
 \label{dcsbm_algo}
\end{algorithm}
\endgroup

\section{Empirical model validation} \label{model_vali}

In order to validate the modelling approach in \eqref{full_model}, a simulation study has been carried out. Additional results on model validation are also reported in Appendices~\ref{asymptotic} and \ref{app_sim}. 

\subsection{Gaussian mixture modelling of DCSBM embeddings} \label{model_sim}

First, a simple simulation is used to show that $k$-means or Gaussian mixtures 
are not 
appropriate for modelling an embedding $\hat{\mvec X}$  
under the DCSBM, even when the row-normalised embedding $\tilde{\mvec X}$ is used.  
The ASE was obtained from a simulated DCSBM  
with $n=\numprint{1000}$ nodes, $K=2$ and $d=2$, with an equal number of nodes allocated to each group, and $B_{11}=0.1, B_{12}=B_{21}=0.05$ and $B_{22}=0.15$, corrected by parameters $\rho_i\sim\text{Beta}(2,1)$. 
Since the community allocations $\vec z$ are known \textit{a priori} in the simulation, it is possible to evaluate the community-specific distributions. 
The results are plotted in Figure~\ref{comparison_coordinates}, which shows the scatterplot and histograms of the marginal distributions for the two-dimensional adjacency spectral embedding $\hat{\mvec X}$ and its normalised version $\tilde{\mvec X}$, and for its transformation to spherical coordinates $\hat{\bm\Theta}$, all labelled by community. 
From Figure~\ref{embedding_X}, it is clear that applying $k$-means or Gaussian mixture modelling on $\hat{\mvec X}$ 
is suboptimal, since the joint distribution or cluster-specific marginal distributions are 
not normally distributed, as predicted by the ASE-CLT \eqref{ase-clt}. Figure~\ref{embedding_X_tilde} shows that row-normalisation is beneficial, since the marginal distributions for $\tilde{\mvec X}_2$ are normally distributed within each community, but it is 
not appropriate for modelling the joint distribution and for at least one of the two marginal distributions for $\tilde{\mvec X}_1$. On the other hand, the transformation $\hat{\bm\Theta}$ in Figure~\ref{embedding_transformed} visually meets the assumption of normality. 
This visual impression is confirmed in Appendix~\ref{asymptotic}, which describes a CLT which strongly supports the normality 
in $\hat{\bm\Theta}$ for the two-dimensional case. 

\begingroup

\begin{figure}[!t]
\centering
\begin{subfigure}[t]{.325\textwidth}
\centering
\caption{Scatterplot of $\hat{\vec x}_i$}
\includegraphics[width=0.945\textwidth]{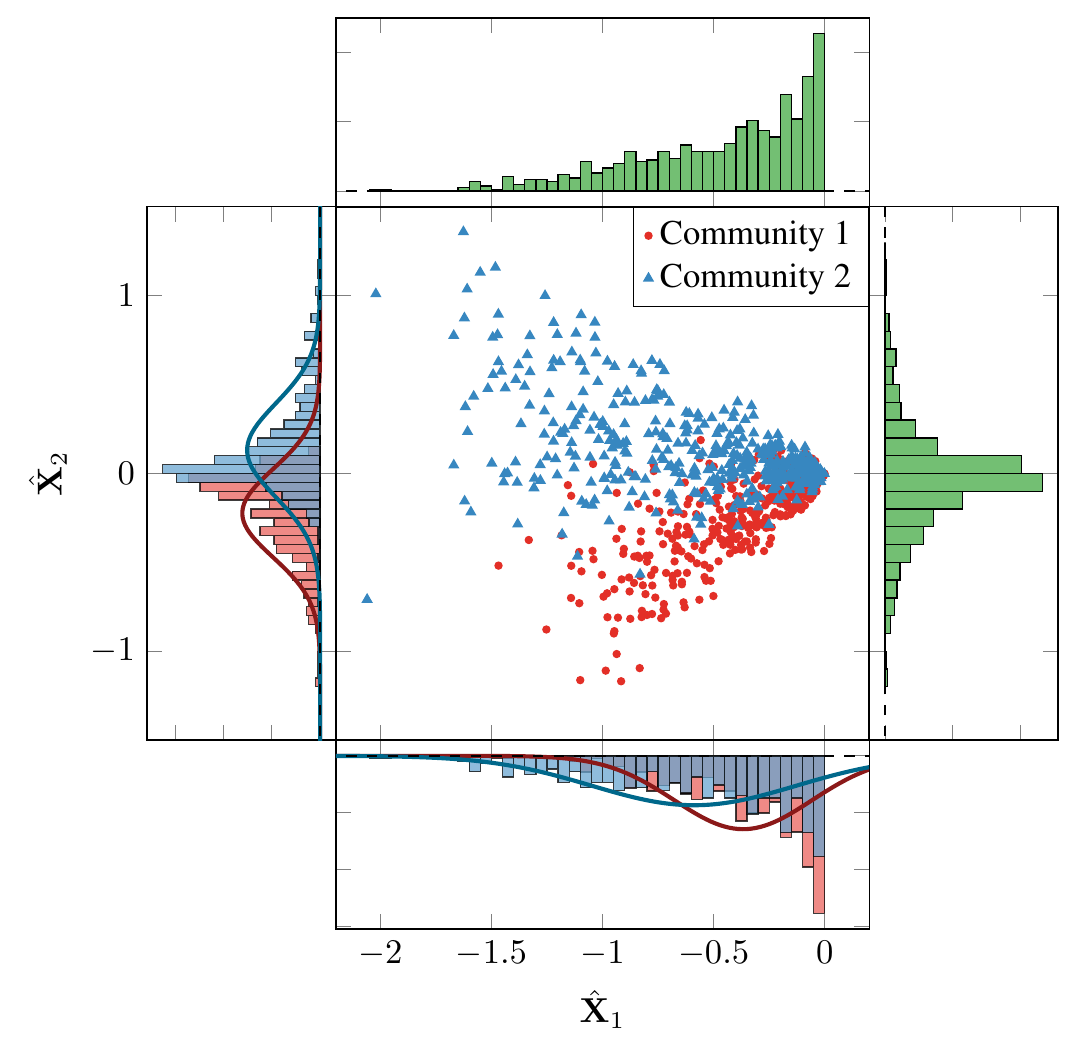}
\label{embedding_X}
\end{subfigure}
\begin{subfigure}[t]{.325\textwidth}
\centering
\caption{Scatterplot of $\tilde{\vec x}_i=\hat{\vec x}_i/\norm{\hat{\vec x}_i}$}
\includegraphics[width=0.97\textwidth]{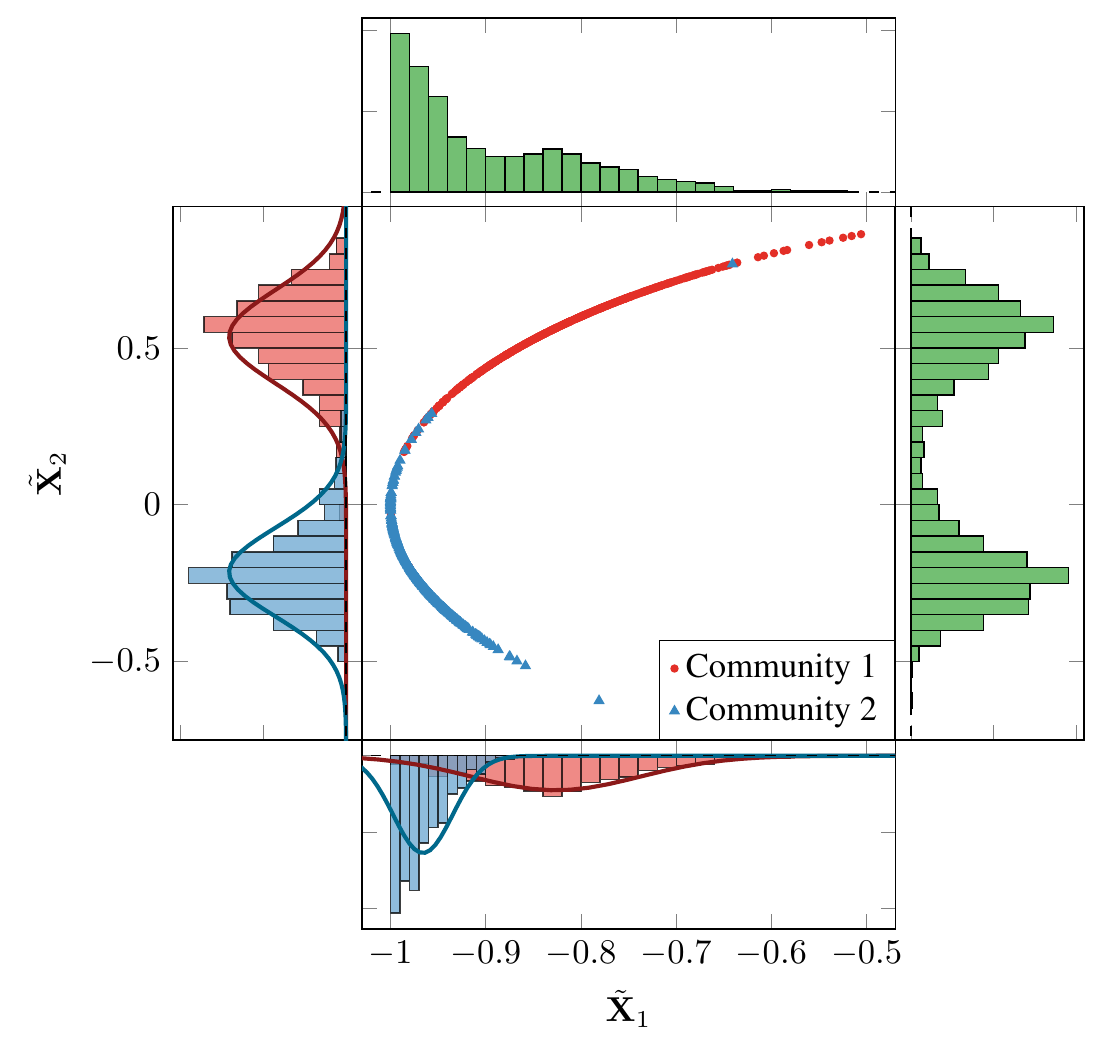}
\label{embedding_X_tilde}
\end{subfigure}
\begin{subfigure}[t]{.325\textwidth}
\centering
\caption{Histograms of $\hat\theta_i=f_2(\hat{\vec x}_i)$}
\includegraphics[width=.975\textwidth]{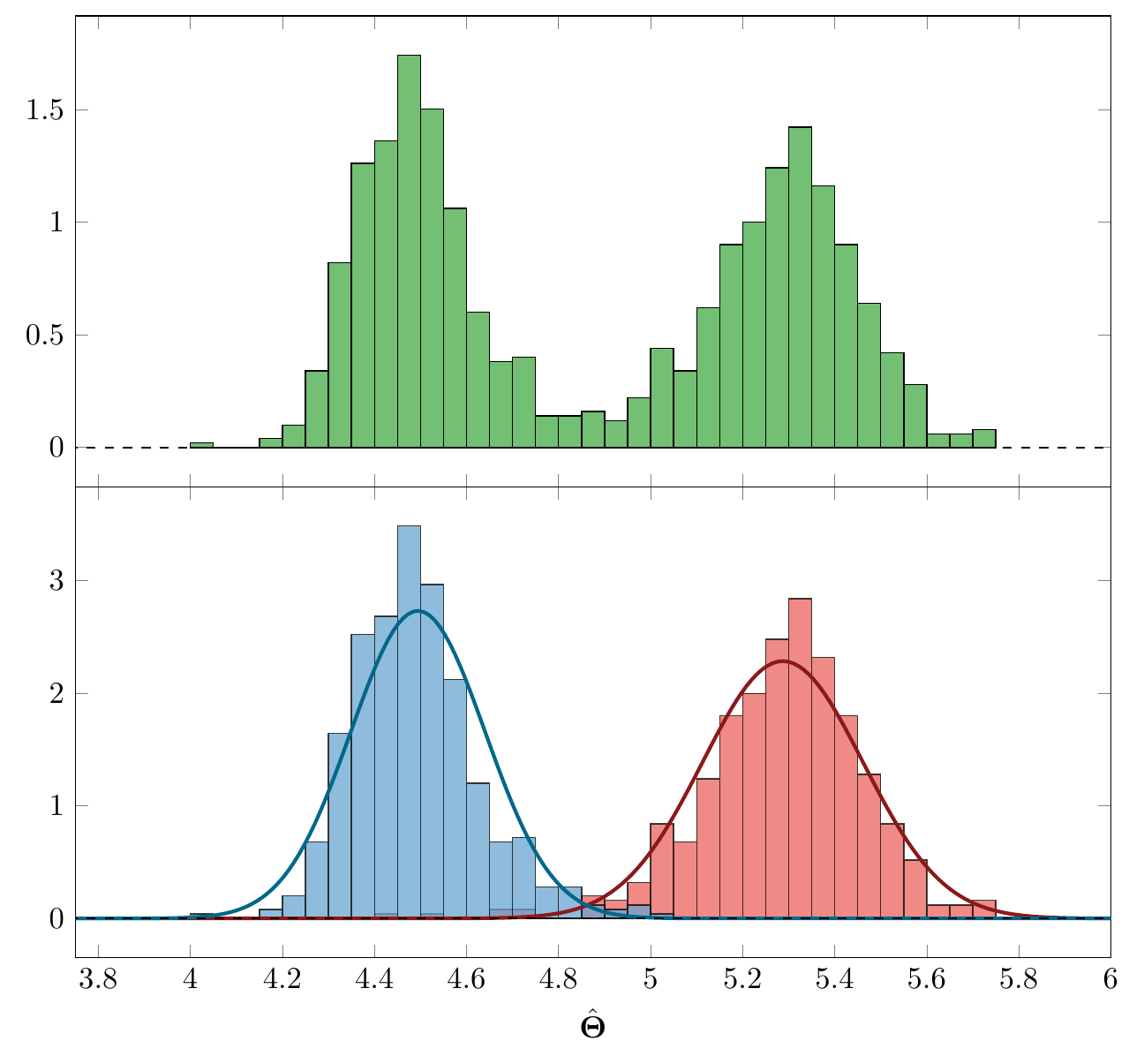}
\label{embedding_transformed}
\end{subfigure}
\caption{Plots of $\hat{\vec x}_i$, $\tilde{\vec x}_i=\hat{\vec x}_i/\norm{\hat{\vec x}_i}$ and $\hat\theta_i=f_2(\hat{\vec x}_i)$, obtained from the two dimensional ASE of a simulated DCSBM. Joint (green) and community-specific (blue and red) marginal distributions with MLE 
Gaussian fit are also shown. 
}
\label{comparison_coordinates}
\end{figure}
\endgroup

\subsection{Structure of the likelihood}
\label{und_dcsbm}

In order to validate the conjecture on the model likelihood proposed in \eqref{full_model}, a simulation study has been carried out. $N=\numprint{1000}$ DCSBMs with $n=\numprint{2000}$ nodes have been simulated, fixing $K=3$, and pre-allocating the nodes to equal-sized clusters. 
The community-specific latent positions used in the simulation are $\bm\mu_1=(0.7,0.4,0.1), \bm\mu_2=(0.1,0.1,0.5)$ and $\bm\mu_3=(0.4,0.8,-0.1)$, resulting in the block-probability matrix:

\begingroup

\begin{equation}
\mvec B = \begin{bmatrix} 0.66 & 0.16 & 0.59 \\ 0.16 & 0.27 & 0.07 \\ 0.59 & 0.07 & 0.81 \end{bmatrix}. \label{blockmatrix}
\end{equation}
\endgroup
The matrix is positive definite and has full rank $3$, implying that $d=2$ in the embedding $\hat{\bm\Theta}$. For each of the $N$ simulations, the link probabilities are corrected using the degree-correction parameters $\rho_i$ sampled from a $\mathrm{Uniform}(0,1)$ distribution, and the adjacency matrices $\mvec A$ are obtained using \eqref{dcsbm}. For each of the simulated graphs ASE is calculated for a large value of $m$. 
The results are summarised in Figure~\ref{simulation_results}.

\begingroup

\begin{figure}[!p]
\centering
\begin{subfigure}[t]{.26\textwidth}
\centering
\caption{Boxplots of $\hat{\bm\vartheta}_{k,:2}$}
\includegraphics[width=.975\textwidth]{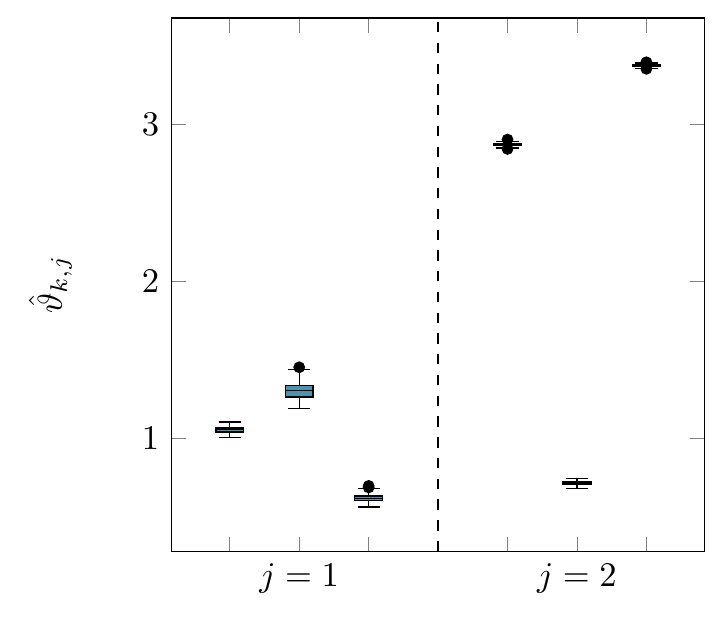}
\label{means}
\end{subfigure}
\begin{subfigure}[t]{.73\textwidth}
\centering
\caption{Boxplots of $\hat{\bm\Sigma}_k$}
\includegraphics[width=.975\textwidth]{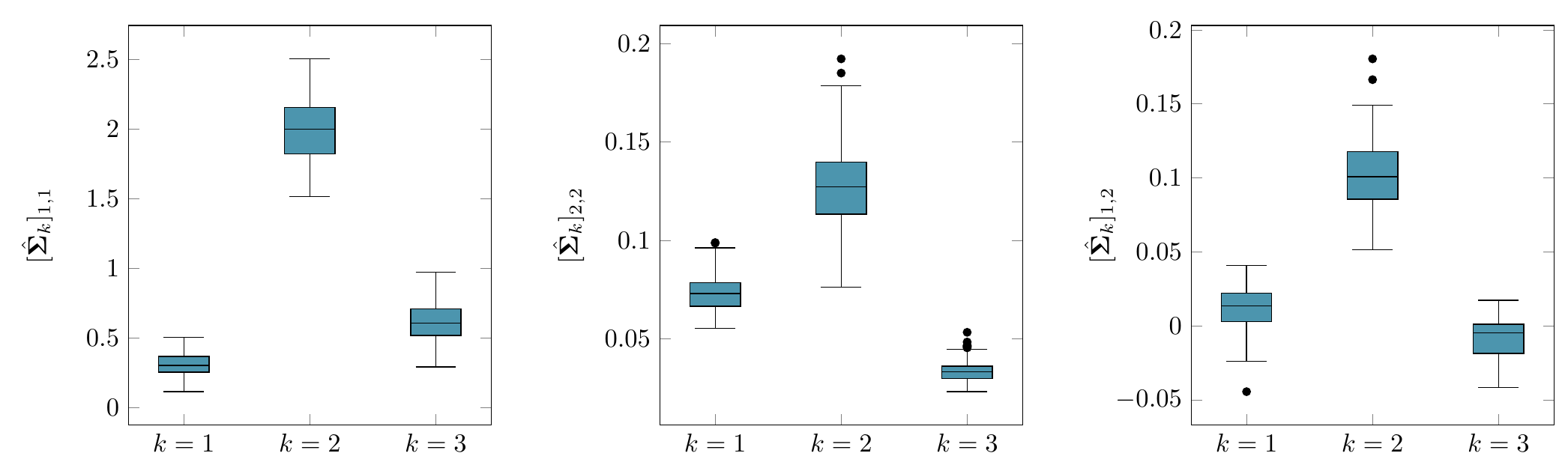}
\label{covs}
\end{subfigure}
\centering
\begin{subfigure}[t]{.325\textwidth}
\centering
\caption{Boxplots of $\hat{\bm\vartheta}_{k,2:}$}
\includegraphics[width=.975\textwidth]{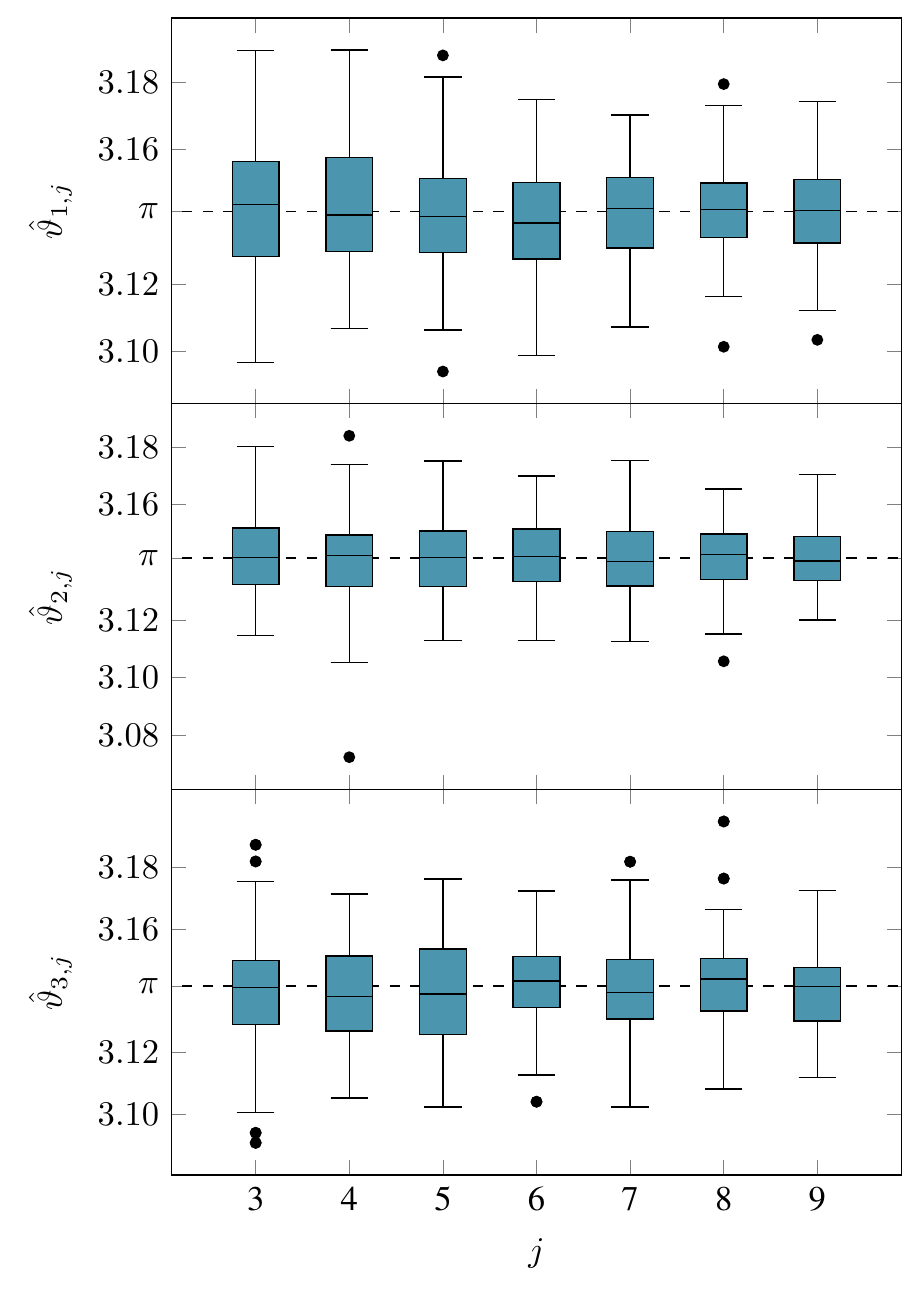}
\label{means_redundant}
\end{subfigure}
\begin{subfigure}[t]{.325\textwidth}
\centering
\caption{Boxplots of $[\hat{\bm\Sigma}_k]_{j,j}$}
\includegraphics[width=.975\textwidth]{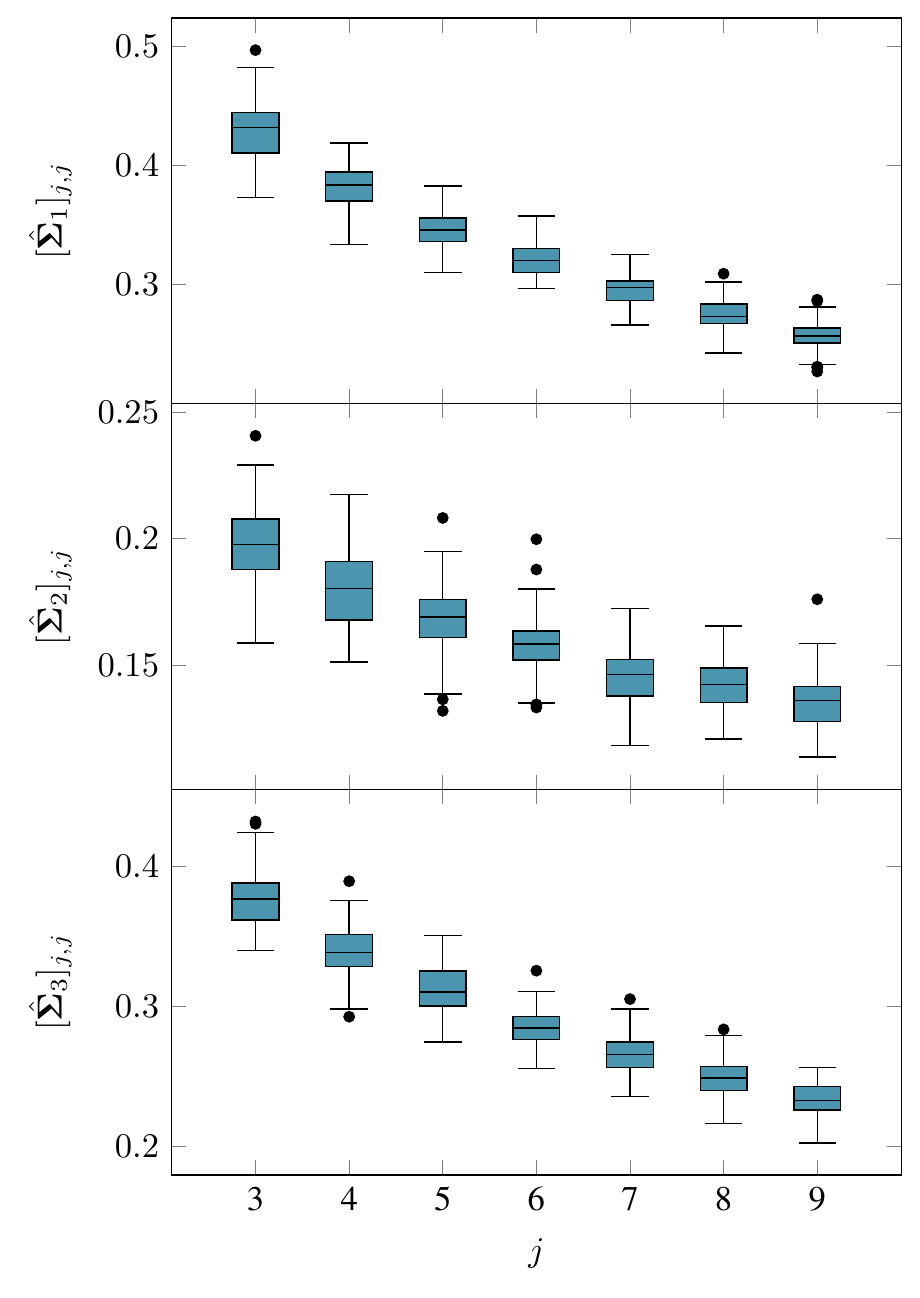}
\label{covs_redundant}
\end{subfigure}
\begin{subfigure}[t]{.325\textwidth}
\centering
\caption{Boxplots of $[\hat{\bm\Sigma}_k]_{1,j}$}
\includegraphics[width=.975\textwidth]{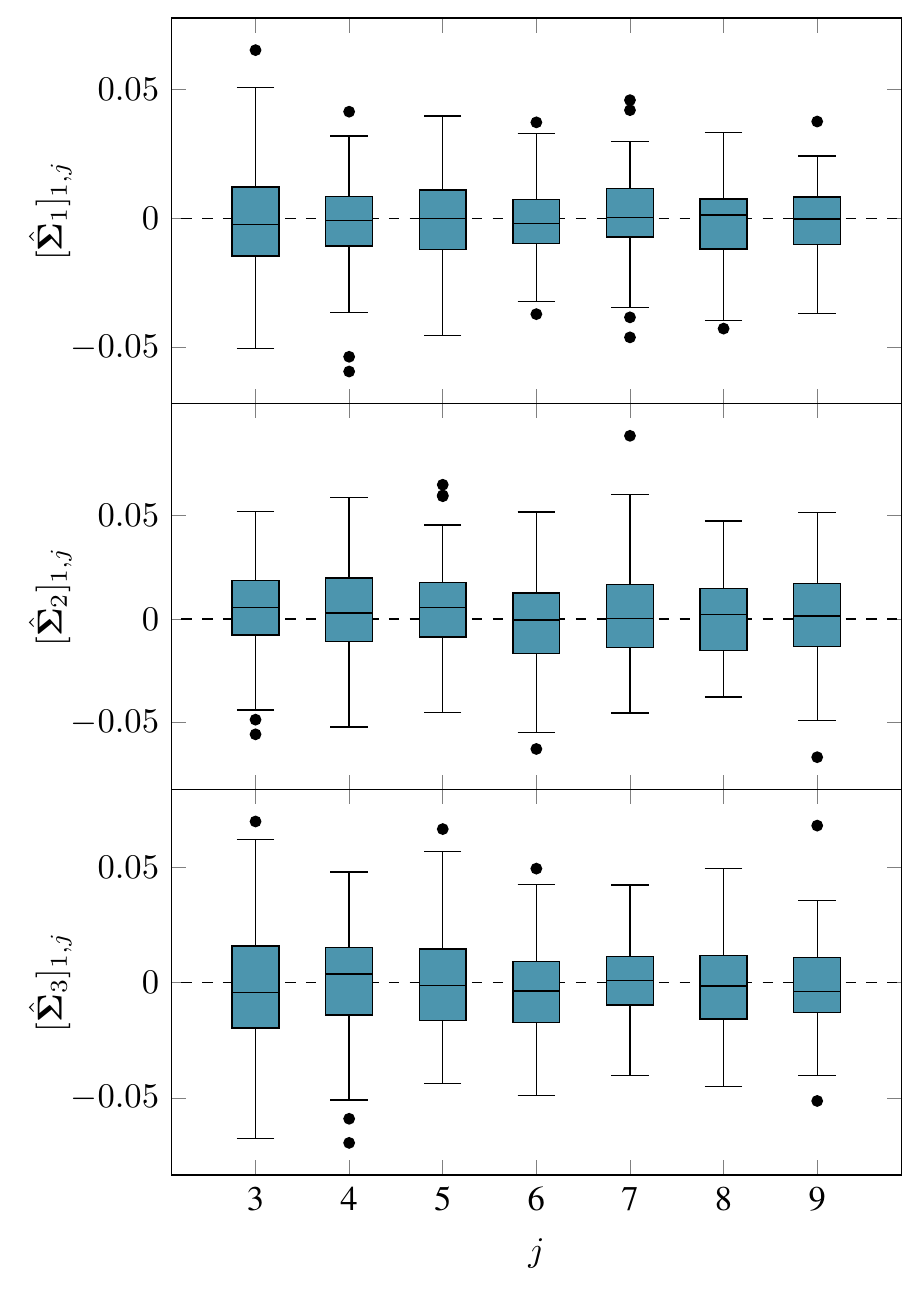}
\label{covs_redundant2}
\end{subfigure}
\begin{subfigure}[t]{0.43\textwidth}
\centering
\caption{Estimated $\hat{\bm\Sigma}_1$}
\includegraphics[width=.975\textwidth]{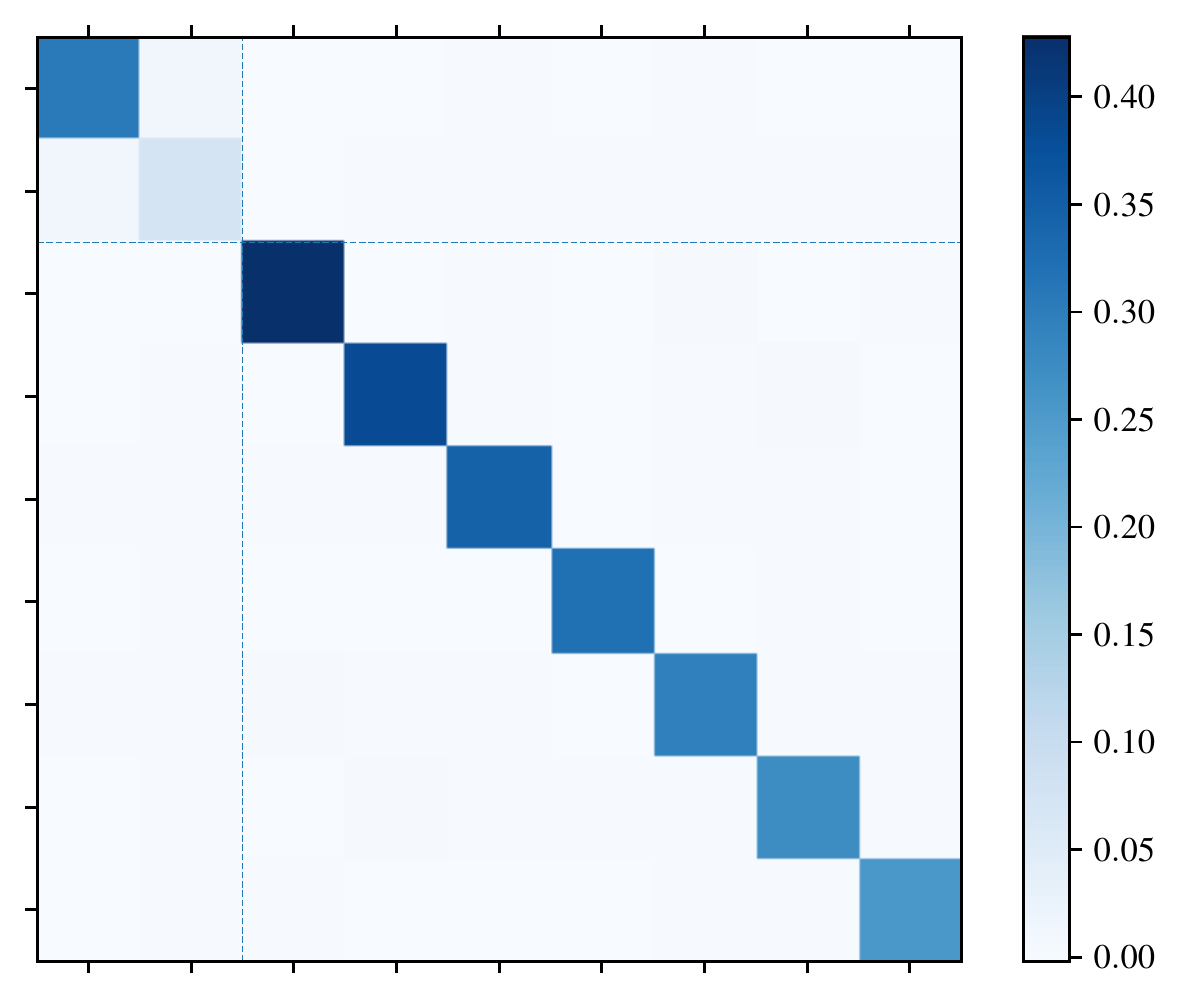}
\label{covs_matrix}
\end{subfigure}
\begin{subfigure}[t]{0.545\textwidth}
\centering
\caption{Boxplots of KS scores for Gaussian fit in $\hat{\bm\Theta}_{d:}$}
\includegraphics[width=.975\textwidth]{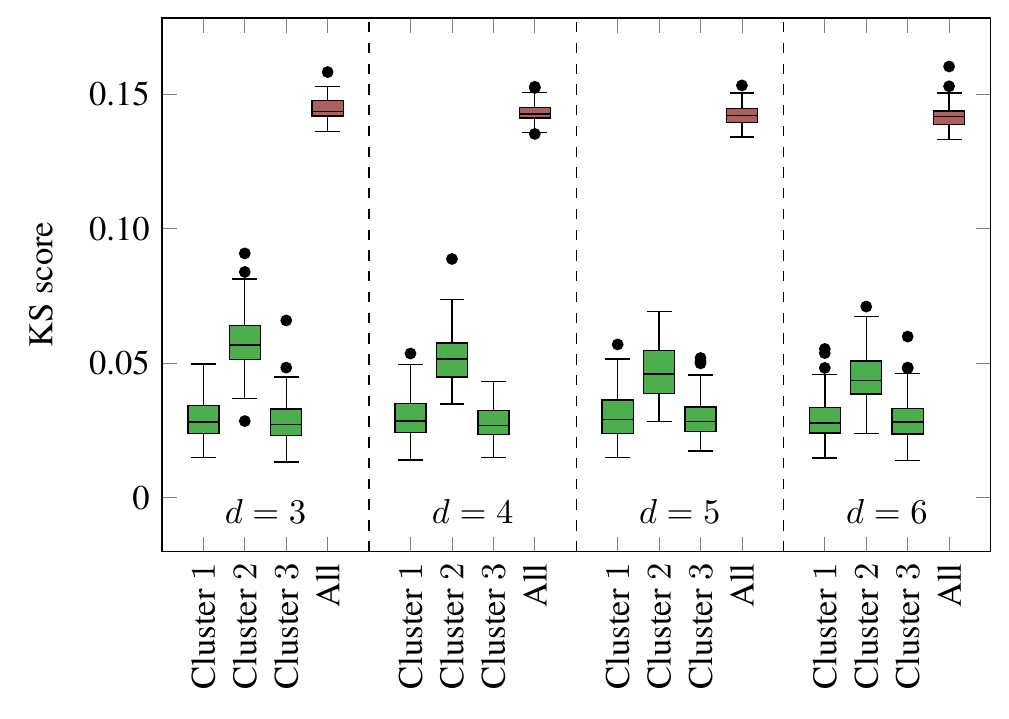}
\label{ks_scores}
\end{subfigure}
\caption{Boxplots for $N=\numprint{1000}$ simulations of a degree-corrected stochastic blockmodel with $n=\numprint{2000}$ nodes, $K=3$, equal number of nodes allocated to each group, and $\mvec B$ described in \eqref{blockmatrix}, corrected by parameters $\rho_i$ sampled from a $\mathrm{Uniform}(0,1)$ distribution.}
\label{simulation_results}
\end{figure}
\endgroup

The true underlying cluster allocations are known in the simulation, and can be used to validate the model assumptions. In this section, the community-specific mean and covariance matrices obtained from the embedding $\hat{\bm\Theta}$ of the sampled graph will be denoted as $\hat{\bm\vartheta}_k$ and $\hat{\bm\Sigma}_k$. Figure~\ref{means} shows the boxplots of the community-specific estimated means for the first two components of $\hat{\bm\Theta}$. The mean values show minimal variation across the different simulations, and are clearly different from $0$ and differ between clusters. In Figure~\ref{covs}, boxplots for the community-specific estimated variances in $\hat{\bm\Theta}_{:d}$ are plotted. Again, it seems that having cluster-specific variances is sensible, and at least one of the covariances is significantly different from $0$, as expected from the theory. On the other hand, Figure~\ref{means_redundant} shows the boxplots for the community-specific estimated means in $\hat{\bm\Theta}_{d:}$, which 
are all centred at $\pi$. 
Hence, the assumption on the mean structure in \eqref{full_model} seems to be justified. 

A potentially more controversial modelling choice is the community-specific variance for the last $m-d$ components of the embedding. Figure~\ref{covs_redundant} shows the estimated variances for each community on different dimensions exceeding $2$. It is clear that the variance is different across different communities. This consideration is reinforced by Figure~\ref{ks_scores}, which shows the boxplots of Kolmogorov-Smirnov (KS) scores obtained from fitting community-specific Gaussian distributions on the dimensions exceeding $d$, compared to the KS score for a Gaussian fit on all the estimated latent positions for the corresponding dimension. Clearly, a correct modelling approach must use community-specific variances. This implies that 
residual cluster information is present also in the last $m-d$ dimensions of the embedding. Finally, Figure~\ref{covs_redundant2} plots the boxplots of estimated covariances between $\hat{\bm\Theta}_1$ and $\hat{\bm\Theta}_{2:}$. Consistent with the model in \eqref{full_model}, the correlations are scattered around $0$, and the assumption of independence between those components seems reasonable. 
This is
confirmed by the plot of the average estimated covariance matrix for the first community, in Figure~\ref{covs_matrix}. 

\subsection{Normality of the spherical coordinates} 
\label{mardia_sec}

The most important comparison is to establish whether the embedding $\hat{\bm\Theta}$ is better suited to GMM than the row-normalised embedding $\tilde{\mvec X}$ traditionally used in the literature. To make this comparison, the $p$-values for the two Mardia tests for multivariate normality \citep{Mardia70} have been calculated for each community-specific distribution, for each simulated graph. The tests are based on multivariate extensions of skewness and kurtosis: assume a sequence of random vectors $\vec x_1,\dots,\vec x_\ell\in\mathbb R^d$, and define the sample mean and sample covariance as $\bar{\vec x}=\sum_{i=1}^\ell \vec x_i/\ell$ and $\mvec S=\sum_{i=1}^\ell (\vec x_i-\bar{\vec x})(\vec x_i-\bar{\vec x})^\intercal/\ell$ respectively. \cite{Mardia70} defines two test statistics for multivariate skewness and kurtosis: 
\begin{gather}
T_S = \frac{1}{6\ell}\sum_{i=1}^\ell \sum_{j=1}^\ell \left[(\vec x_i-\bar{\vec x})^\intercal\mvec S^{-1}(\vec x_j-\bar{\vec x})\right]^3, 
\\
T_K = \sqrt{\frac{\ell}{8d(d+2)}}\left\{\frac{1}{\ell}\sum_{i=1}^\ell \left[(\vec x_i-\bar{\vec x})^\intercal\mvec S^{-1}(\vec x_i-\bar{\vec x})\right]^2 - \frac{d(d+2)(\ell-1)}{\ell+1}\right\}. 
\end{gather}
Under the null hypothesis of multivariate normality, 
$T_S\to \chi^2\{d(d+1)(d+2)/6\}$ and $T_K\to \mathbb N(0,1)$ in distribution for $\ell\to\infty$ \citep{Mardia70}. Given observed values of $T_S$ and $T_K$, $p$-values $p_S$ and $p_K$ can be calculated from the asymptotic distribution. 

Under the same setup as the 
simulation in Section~\ref{und_dcsbm}, $p$-values are calculated for the two Mardia tests applied for each community on the spherical embedding $\hat{\bm\Theta}_{:2}$ and the row-normalised embedding $\tilde{\mvec X}_{:3}$. 
Then, binomial sign tests for paired observations are calculated on the differences between the $p$-values obtained from $\hat{\bm\Theta}_{:2}$, and those obtained from $\tilde{\mvec X}_{:3}$, separately for $p_S$ and $p_K$, under the null hypothesis that those are sampled from the same distribution. The alternative hypothesis is that the distribution of the $p$-values obtained from $\hat{\bm\Theta}_{:2}$ is stochastically larger than the corresponding distribution for $\tilde{\mvec X}_{:3}$. The $p$-value of the sign test is $<10^{-10}$ for both skewness and kurtosis, 
confirming the impression in Figure~\ref{comparison_coordinates} that the transformation \eqref{cart_to_sphere} to $\hat{\bm\Theta}$ tends to \textit{Gaussianise} the embeddings $\hat{\mvec X}$ and $\tilde{\mvec X}$. 

\section{Applications and results} \label{results_section}

In this section, the model selection procedure is assessed on simulated DCSBMs and real-world bipartite graphs obtained from the network flow data collected at Imperial College London. 
The DCSBM for bipartite graphs 
is a simple extension of the undirected model, and has a similar RDPG-structure to the bipartite stochastic co-blockmodel \citep[ScBM,][]{Rohe16}. 
In bipartite graphs, the nodes are divided into two non-overlapping groups $V$ and $V^\prime$ such that $E\subseteq V\times V^\prime$. Such networks are conveniently represented by rectangular adjacency matrices $\mvec A\in\{0,1\}^{n\times n^\prime}$, where $n=\abs{V}$ and $n^\prime=\abs{V^\prime}$. Suppose the nodes in $V$ and $V^\prime$ respectively belong to $K$ and $K^\prime$ communities, with respective community allocations ${\vec z} \in\{1,\dots,K\}^n$ and ${\vec z^\prime} \in\{1,\dots,K^\prime\}^{n^\prime}$. Also, suppose for each of the two sets of communities there are latent positions $\bm\mu_k \in \mathbb{R}^d,\ k\in\{1,\dots,K\}$, and $\bm\mu^\prime_\ell \in \mathbb{R}^d,\ \ell\in\{1,\dots,K^\prime\}$, such that $\bm\mu_k^\intercal\bm\mu^\prime_\ell\in[0,1]$. 
This gives the link probability
\begin{equation}
A_{ij}\sim\mathrm{Bernoulli}(\rho_i\rho_j^\prime\bm\mu_{z_i}^\intercal\bm\mu^\prime_{z_j^\prime}), i\in V, j\in V^\prime, \label{coblock}
\end{equation}
where $\rho_i\in[0,1]$ and $\rho_j^\prime\in[0,1]$ are degree correction parameters for each of the nodes in $V$ and $V^\prime$. From $\mvec A$, it is possible to obtain embeddings $\hat{\mvec X}$ and $\hat{\mvec X}^\prime$ using the DASE in Section~\ref{defi_section}, and cluster the two embeddings jointly or separately.
In this work, the quality of the clustering is evaluated using the adjusted Rand index \citep[ARI,][]{Hubert85}. Higher values of the ARI correspond to better clustering performance, reaching a maximum of $1$ for perfect agreement between the estimated clustering and the true labels. 

\subsection{Synthetic networks} 

The performance of the model selection procedure described in Section~\ref{model_selection} is evaluated on simulated DCSBMs. $N=250$ undirected graphs with $n=\numprint{1000}$ and $K\in\{2,3\}$ were simulated, randomly selecting $\mvec B$ from $\mathrm{Uniform}(0,1)^{K\times K}$ and sampling the degree correction parameters from $\mathrm{Beta}(2,1)$. The nodes were allocated to communities of equal size. For each of the graphs, the models of \cite{SannaPassino19} and \cite{Yang19} are applied to the ASE $\hat{\mvec X}$ and its row-normalised version $\tilde{\mvec X}$ for $m=10$, selecting the estimates of $d$ and $K$ using BIC. The value $m=10$ is usually approximately equal in the simulations to the third elbow of the scree plot using the criterion of \cite{Zhu06}, considering a total of $25$ eigenvalues or singular values. Also, the model in \eqref{full_model} is fitted to $\hat{\bm\Theta}$, estimating $d$ and $K$ using the selection procedure 
in Section~\ref{model_selection}, with $K^\ast=6$. 
The results of the simulations are reported in Table~\ref{table_und}. 

A similar simulation has been repeated for bipartite DCScBMs. $N=250$ graphs with $n=\numprint{1000}$ and $n^\prime=\numprint{1500}$ were generated, setting $K=2,\ K^\prime=3$, communities of equal size, $\mvec B\sim\mathrm{Uniform}(0,1)^{K\times K^\prime}$, and $\rho_i\sim\mathrm{Beta}(2,1)$. 
The results are reported in Table~\ref{table_bip}. 

\begingroup

\begin{table}[!t]
\centering
\begin{subtable}{\textwidth}
\centering
\caption{Undirected DCSBM}
\label{table_und}
\begin{tabular}{ c | ccc | ccc }
\toprule
& \multicolumn{3}{c|}{$K=2$} & \multicolumn{3}{c}{$K=3$} \\ 
\midrule
 & $\hat{\mvec X}$ & $\tilde{\mvec X}$ & $\hat{\bm\Theta}$  & $\hat{\mvec X}$  & $\tilde{\mvec X}$ & $\hat{\bm\Theta}$ \\
\midrule
Proportion of correct $d$ & $0.788$ & $0.796$ & $0.972$ & $0.740$ & $0.736$ & $0.748$ \\
Proportion of correct $K$ & $0.000$ & $0.080$ & $0.624$ & $0.000$ & $0.092$ & $0.296$ \\
Average ARI & $0.339$ & $0.510$ & $0.764$ & $0.594$ & $0.748$ & $0.858$ \\
\bottomrule
\end{tabular}
\end{subtable}
\begin{subtable}{\textwidth}
\vspace{1em}
\centering
\caption{Bipartite DCScBM}
\label{table_bip}
\begin{tabular}{ c | ccc | ccc}
\toprule
& \multicolumn{3}{c|}{$K=2$} & \multicolumn{3}{c}{$K^\prime=3$} \\ 
\midrule
 & $\hat{\mvec X}$ & $\tilde{\mvec X}$ & $\hat{\bm\Theta}$  & $\hat{\mvec X}^\prime$      & $\tilde{\mvec X}^\prime$ & $\hat{\bm\Theta}^\prime$ \\
\midrule
Proportion of correct $d$ & $0.940$ & $0.948$ & $0.876$ & $0.888$ & $0.916$ & $0.896$ \\
Proportion of correct $K$ & $0.000$ & $0.056$ & $0.580$ & $0.000$ & $0.096$ & $0.540$\\
Average ARI & $0.374$ & $0.564$ & $0.885$ & $0.490$ & $0.572$ & $0.715$\\
\bottomrule
\end{tabular}
\end{subtable}
\caption{Estimated performance for $N=250$ simulated DCSBMs and bipartite DCScBMs.}
\label{table_est}
\end{table}
\endgroup

The table shows that the 
transformed embedding sometimes has a slightly inferior performance when estimating the correct value of the latent dimension $d$ (\textit{cf.} Table~\ref{table_bip}), but 
outperforms the alternative methodologies significantly 
in the ability to estimate the 
number of communities $K$. In particular, the Gaussian mixture model is not well suited to either the standard embedding $\hat{\mvec X}$ nor the row-normalised $\tilde{\mvec X}$, and the distortion caused by the degree-corrections and row-normalisation does not allow correct estimation of $K$. This problem is alleviated when the spherical coordinates estimator is used. The improvement is reflected in a significant difference in the clustering performance, demonstrated by the average ARI scores for the three different procedures. The table also shows that estimates of $d$ based on the model of \cite{SannaPassino19} and \cite{Yang19} on $\hat{\mvec X}$ and $\hat{\mvec X}^\prime$ seem to be slightly more accurate than alternative methods on the DCScBM. It might be therefore tempting to construct a hybrid model that uses $\hat{\mvec\Theta}_{:d}$ for the top-$d$ embeddings and $\hat{\mvec X}_{d:}$ for the remaining components, and proceed to select the most appropriate $d$ under such a joint model. Unfortunately, model comparison via BIC is not possible in  
that 
setting. 

\begingroup

\begin{figure}[!t]
\centering
\begin{subfigure}[t]{.32\textwidth}
\centering
\caption{Proportion of correct $d$ \\ Undirected DCSBM}
\includegraphics[width=.9\textwidth]{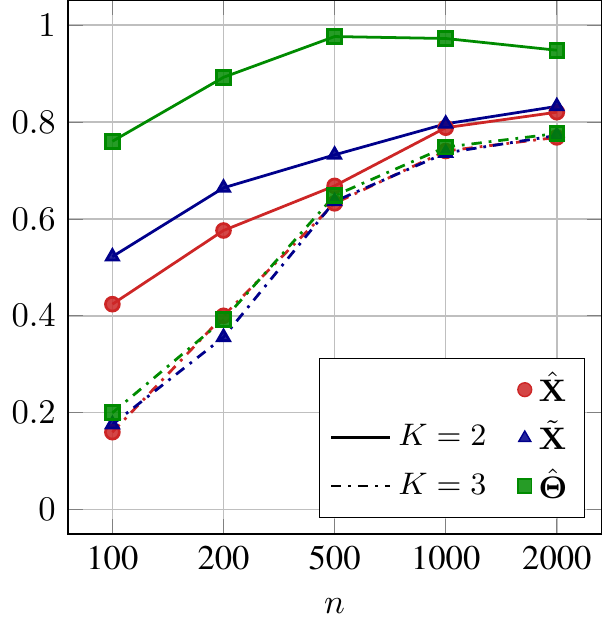}
\label{d_asy}
\end{subfigure}
\begin{subfigure}[t]{.32\textwidth}
\centering
\caption{Average ARI \\ Undirected DCSBM}
\includegraphics[width=.9\textwidth]{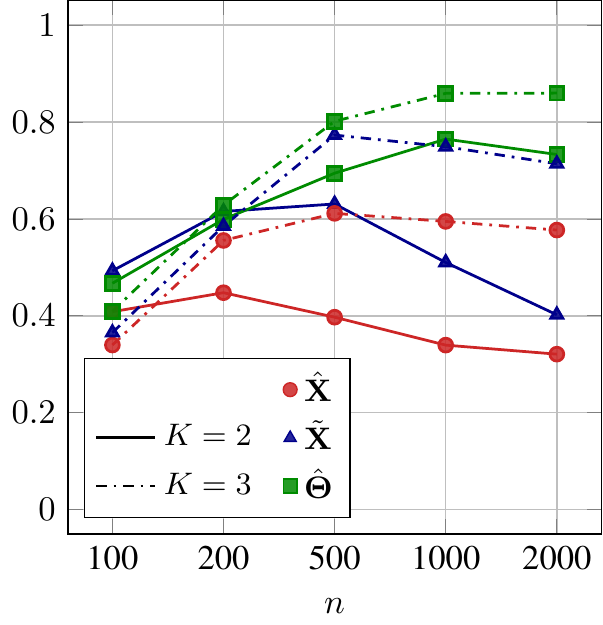}
\label{ari_asy}
\end{subfigure}
\begin{subfigure}[t]{.32\textwidth}
\centering
\caption{Average ARI \\ Bipartite DCScBM}
\includegraphics[width=0.975\textwidth]{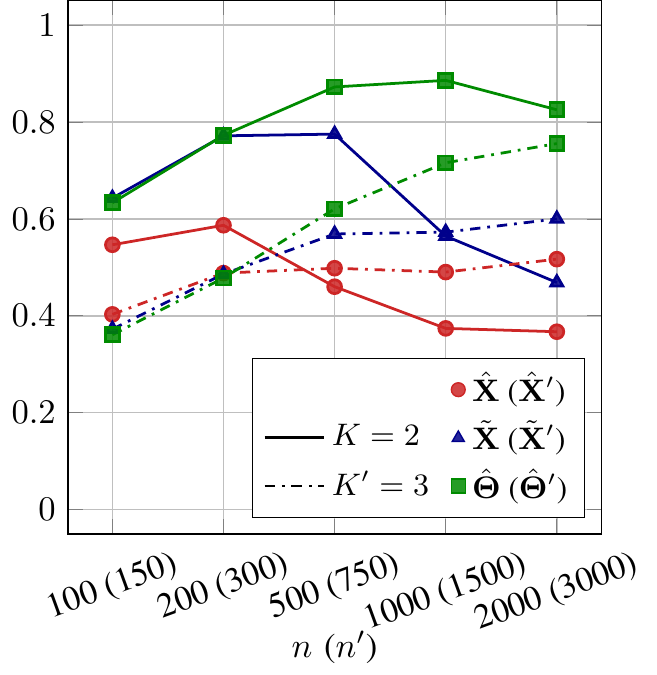}
\label{rec_asy}
\end{subfigure}
\caption{Estimated performance for $N=250$ simulated DCSBMs and DCScBMs, for $n\in\{100,200,500,1000,2000\}$. For bipartite DCScBMs, $n^\prime\in\{150,300,750,1500,3000\}$.} 
\label{asy_sim}
\end{figure}
\endgroup

The simulation is also repeated for different values of $n$, evaluating the asymptotic behaviour of the proposed community detection procedure. The results are plotted in Figure~\ref{asy_sim}, demonstrating that the performance of the spherical coordinates estimator improves when the number of nodes in the graph increases. This appears to be consistent with the CLT presented in Appendix~\ref{asymptotic}. On the other hand, the results obtained from the alternative estimators degrade with $n$, providing further evidence that the proposed model \eqref{full_model} appears to be more appropriate for community detection under the DCSBM.

Also, the boxplots for the paired differences between ARIs are plotted in Figure~\ref{boxplots}. The clustering based on $\hat{\bm\Theta}$ consistently outperforms $\hat{\mvec X}$ and $\tilde{\mvec X}$ 
(\textit{cf.} Figure~\ref{box_all} and~\ref{box_differences_bipartite}). The difference can be quantitatively evaluated using binomial sign tests for paired observations, similarly to Section~\ref{mardia_sec}. For both undirected and bipartite graphs, the $p$-values of the sign tests are $<10^{-10}$, overwhelmingly suggesting that the clustering based on $\hat{\bm\Theta}$ is superior to the competing methodologies. Furthermore, the difference increases when the number of nodes increases (\textit{cf.} Figure~\ref{box_differences}). Overall, the simulations suggest that the proposed spectral clustering procedure for estimation of the DCSBM, based on the spherical coordinates of the ASE estimator of the latent positions, appears to outperform competing estimators, including spectral clustering on the row-normalised ASE estimator. 

\begingroup

\begin{figure}[!t]
\centering
\begin{subfigure}[t]{.32\textwidth}
\centering
\caption{Undirected graph \\ $K\in\{2,3\}$, $n=1000$}
\includegraphics[height=.875\textwidth]{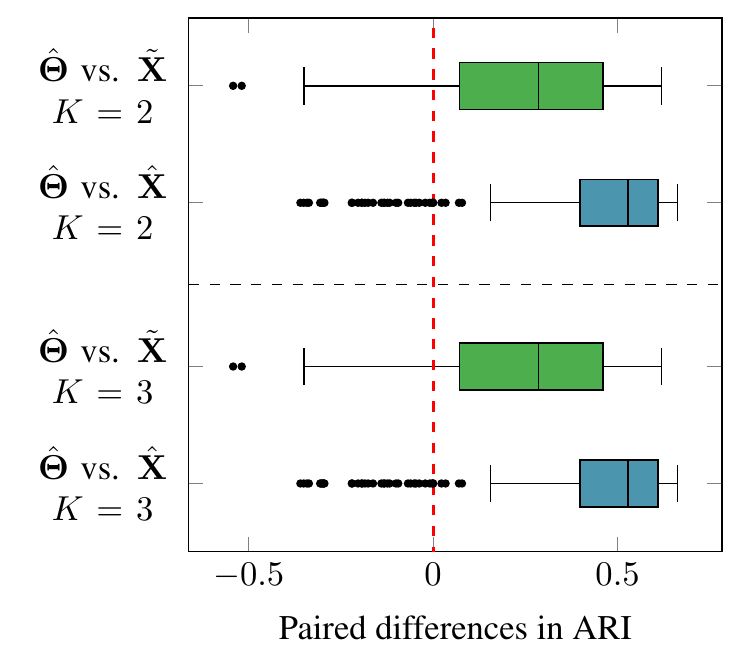}
\label{box_all}
\end{subfigure}
\begin{subfigure}[t]{.32\textwidth}
\centering
\caption{Undirected graph, $K=2$ \\ Varying $n$, $\hat{\bm\Theta}$ vs. $\tilde{\mvec X}$}
\includegraphics[height=.875\textwidth]{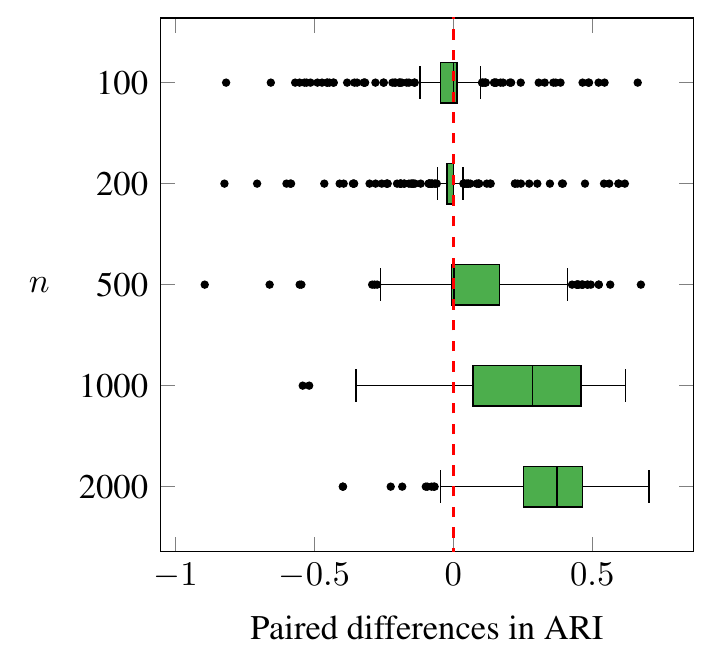}
\label{box_differences}
\end{subfigure}
\begin{subfigure}[t]{.32\textwidth}
\centering
\caption{Bipartite graph, $K=2$, $K^\prime=3$, $n=1000$, $n^\prime=1500$}
\includegraphics[height=.875\textwidth]{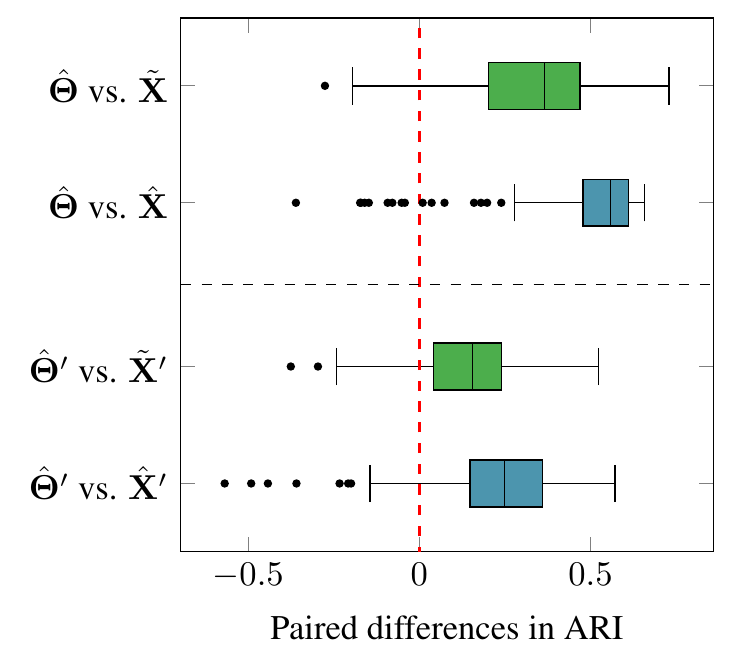}
\label{box_differences_bipartite}
\end{subfigure}
\caption{Boxplots of differences in ARI for $N=250$ simulated DCSBMs and DCScBM.} 
\label{boxplots}
\end{figure}
\endgroup

\subsection{Imperial College network flow data}  \label{results_icl}

The model proposed in this work has been 
specifically developed for clustering 
networks obtained from the network flow data collected at Imperial College London (ICL). 
Finding communities of machines with similar behaviour is important in network monitoring and intrusion detection \citep{Neil13}.
In this application, the edges relate to all the HTTP (port 80) and HTTPS (port 443) connections observed from machines hosted in computer labs in different departments at ICL in January 2020. The source nodes $V$ are computers hosted in college laboratories, and the destination nodes $V^\prime$ are internet servers. 
Intuitively, computers in a laboratory tend to have a heterogeneous degree distribution of their web connections because they are not used uniformly: a computer located closed to the entrance of the laboratory might be used more than a machine at the back. Therefore, the total time of activity is different across machines in the same community, which suggests that the DCSBM is an appropriate modelling choice.   
This is confirmed by the within-community degree distributions in Figure~\ref{sim_sbms} (Section~\ref{intro_section}), which compares one of the Imperial College networks (ICL2, \textit{cf.} Figure~\ref{icl_degree}) with a simulated SBM and DCSBM (\textit{cf.} Figures~\ref{sbm_degree} and~\ref{dcsbm_degree})\footnote{Figure~\ref{sbm_degree} displays the within-community out-degree distribution of a simulated ScBM, see \eqref{coblock}, with $K=K^\prime=4$, equal-sized communities, and block-community matrix $\mvec B/2$ (see below). Figure~\ref{dcsbm_degree} displays the same distribution for a simulated DCScBM with block-community matrix $2\mvec B$, corrected sampling $\rho_i,\rho_j^\prime\sim\mathrm{Beta}(3,5)$. For $\mvec B$: $B_{11}=0.35$, $B_{22}=0.25$, $B_{33}=0.15$, $B_{44}=0.1$, $B_{k\ell}=0.1$ if $k\neq\ell$.}.

Three real-world computer networks are considered here,  
and corresponding summary statistics are presented in Table~\ref{icl_networks}, where ZG$\ell$ denotes the position of the $\ell$-th elbow of the scree plot according to the method of \cite{Zhu06}. 
For the source nodes,  
a known underlying community structure is given by the department 
to which each machine belongs.  
ICL1 corresponds to machines hosted in the departments of Physics, Electrical Engineering, and Earth Science, whereas computers in Chemistry, Civil Engineering, Mathematics, and Medicine are considered in ICL2. For ICL3, the departments are Aeronautical Engineering, Civil Engineering, Electrical Engineering, Mathematics, and Physics.
Students use the computer laboratories for tutorials and classes, and therefore some variation might be expected in the activities of different machines across different departments.

\begingroup

\begin{table}[!t]
\centering
\begin{tabular}{c | ccc | c | cc}
\toprule
Name & $\abs{V}$ & $\abs{V^\prime}$ & $\abs{E}$ & $K$ & ZG3 & ZG4 \\ 
\midrule
ICL1 & $\numprint{628}$ & $\numprint{54111}$ & $\numprint{668155}$ & 3 & 24 & 52 \\
ICL2 & $\numprint{439}$ & $\numprint{60635}$ & $\numprint{717912}$ & 4 & 24 & 53 \\ 
ICL3 & $\numprint{1011}$ & $\numprint{84664}$ & $\numprint{1470074}$ & 5 & 27 & 51 \\ 
\bottomrule
\end{tabular}
\caption{Summary statistics for the Imperial College London computer networks.}
\label{icl_networks}
\end{table}
\endgroup

\begingroup

\begin{figure}[!t]
\centering
\begin{subfigure}[t]{.325\textwidth}
\centering
\caption{$\hat{\mvec X}_{:2}$}
\includegraphics[width=.95\textwidth]{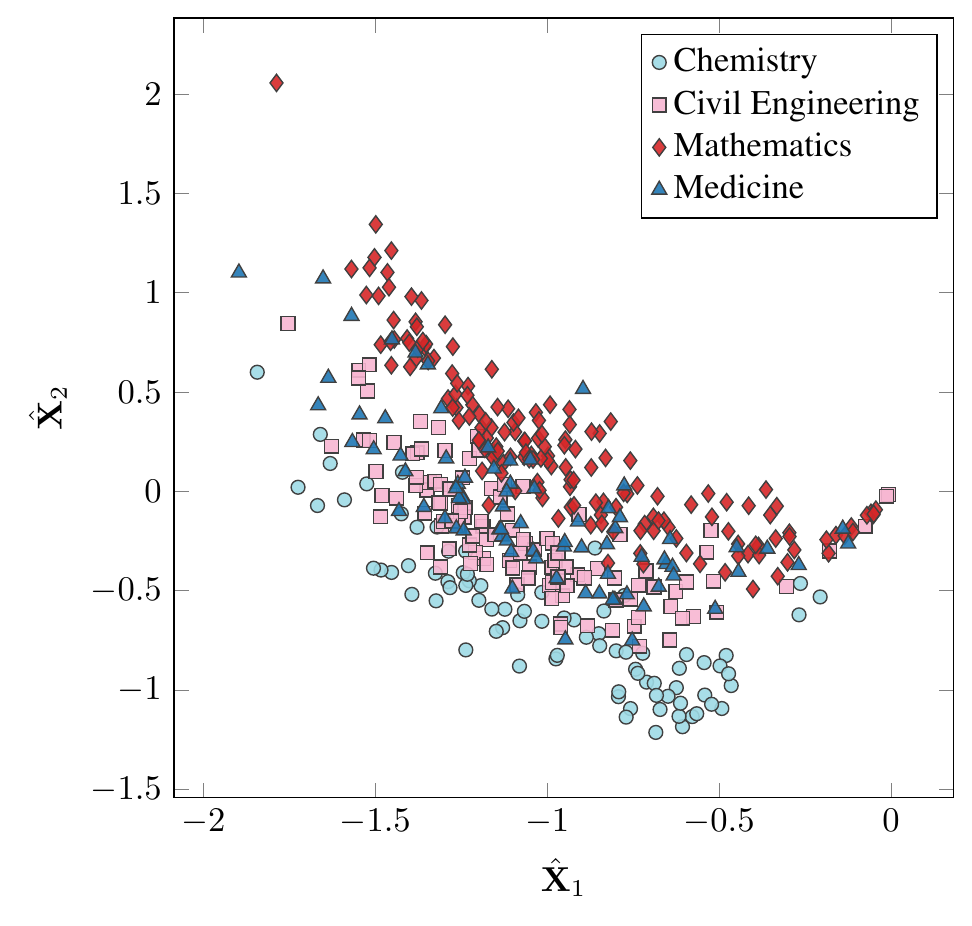}
\label{icl_X}
\end{subfigure}
\begin{subfigure}[t]{.325\textwidth}
\centering
\caption{$\tilde{\mvec X}_{:2}$}
\includegraphics[width=.95\textwidth]{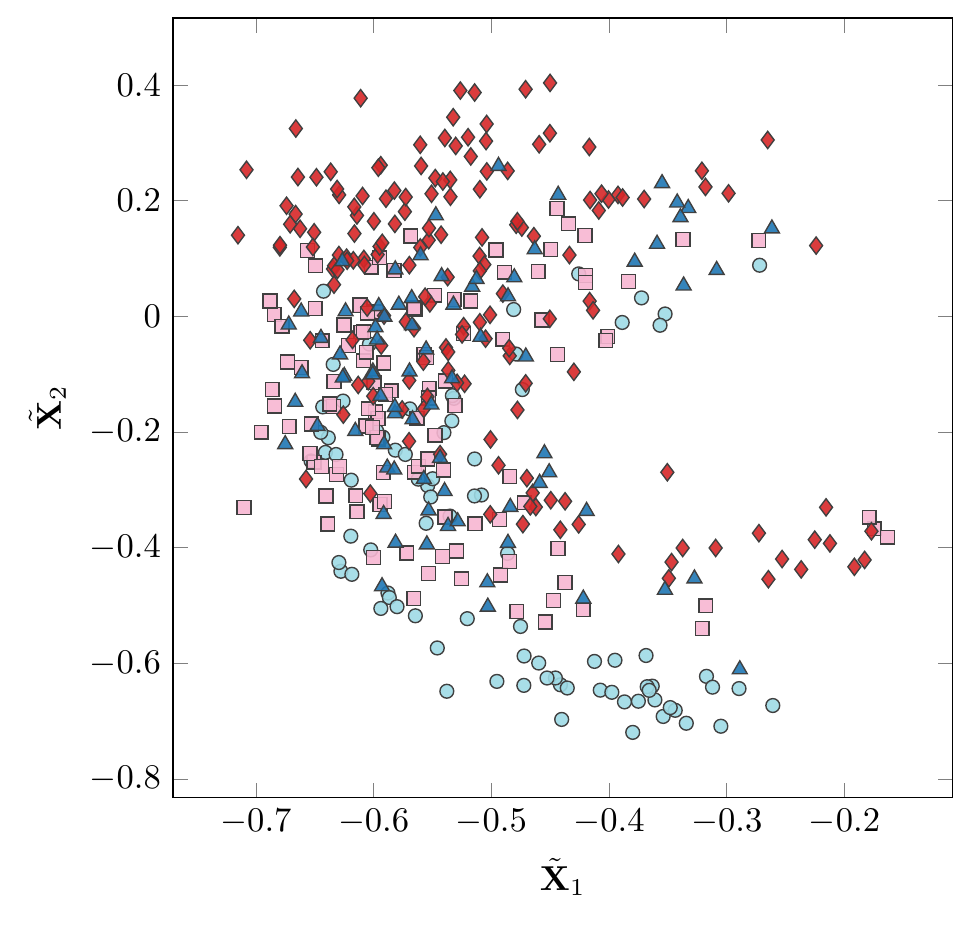}
\label{icl_X_tilde}
\end{subfigure}
\begin{subfigure}[t]{.325\textwidth}
\centering
\caption{$\hat{\bm\Theta}_{:2}$}
\includegraphics[width=.95\textwidth]{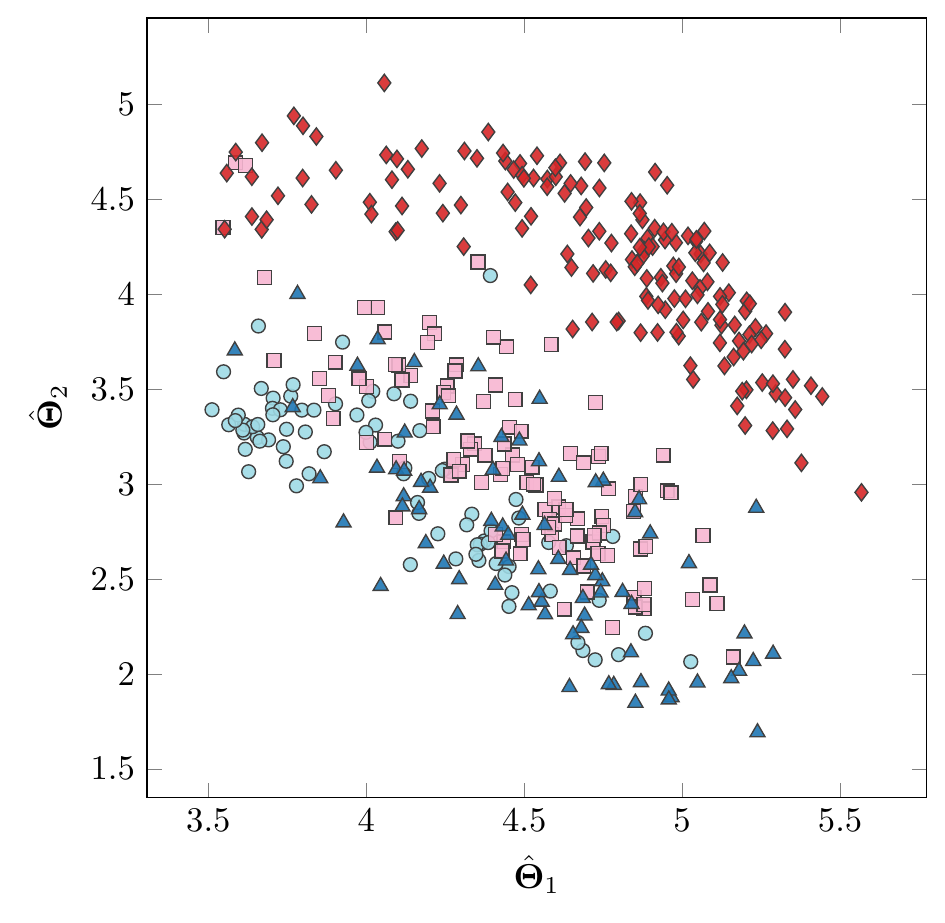}
\label{icl_theta}
\end{subfigure} 
\caption{ICL2: scatterplot of the leading two dimensions for $\hat{\mvec X}$, $\tilde{\mvec X}$ and $\hat{\bm\Theta}$.}
\label{icl_scatterplots}
\end{figure}
\endgroup

Figure~\ref{icl_scatterplots} shows the scatterplots of the leading 2 dimensions of the $m$-dimensional embeddings $\hat{\mvec X}$, $\tilde{\mvec X}$ and $\hat{\bm\Theta}$ for $m=30$ for ICL2, showing that the clustering task is particularly difficult in this network, and there is not much separation between the communities. Despite this, the transformation to spherical coordinates (\textit{cf.} Figure~\ref{icl_theta}) appears to make the communities more Gaussian-shaped, as opposed to the standard and row-normalised DASE, where the within-community embeddings are curved (\textit{cf.} Figure~\ref{icl_X} and \ref{icl_X_tilde}), and therefore not amenable to Gaussian mixture modelling. This impression is quantitatively confirmed by the Mardia tests for multivariate normality on each community (\textit{cf.} Section~\ref{mardia_sec}): the difference of the log-$p$-values obtained from the spherical coordinates transformation and the standard DASE is on average $\approx 42$ for the kurtosis and $\approx82$ for the skewness, further demonstrating that the proposed transformation tends to \textit{Gaussianise} the embedding. 
 
For each of the three ICL network graphs, the DASE has been calculated for $m=30$ and $m=50$, and the model in \cite{SannaPassino19} and \cite{Yang19} has been fitted on the source embeddings $\hat{\mvec X}$ and the row-normalised version $\tilde{\mvec X}$, whereas the model \eqref{full_model} is used on the transformation $\hat{\bm\Theta}$, setting $K^\ast=20$. The resulting estimated values of $d$ and $K$, obtained using the minimum BIC, are reported in Table~\ref{table_bic}. In order to reduce the sensitivity to initialisation, the model was fitted $10$ times for each pair $(d,K)$, and the parameter estimates corresponding to the minimum BIC were retained. The choice of $m\in\{30,50\}$ corresponds to values around the third and fourth elbows in the scree plot of singular values, according to the criterion of \cite{Zhu06} (\textit{cf.} Table~\ref{icl_networks}).   

\begingroup

\begin{table}[!t]
\centering
\begin{subtable}{\textwidth}
\centering
\caption{Estimated $(d,K)$}
\label{table_bic}
\begin{tabular}{ c | ccc | ccc}
\toprule
& \multicolumn{3}{c|}{$m=30$} & \multicolumn{3}{c}{$m=50$} \\
\midrule
 & $\hat{\mvec X}$ & $\tilde{\mvec X}$ & $\hat{\bm\Theta}$ & $\hat{\mvec X}$ & $\tilde{\mvec X}$ & $\hat{\bm\Theta}$ \\
\midrule
ICL1    & $(26,6)$ & $(11,6)$    & $(11,4)$  & $(22,5)$ & $(11,6)$    & $(11,4)$ \\
ICL2    & $(28,5)$ & $(8,7)$    & $(15,4)$ & $(29,4)$ & $(8,7)$    & $(15,4)$ \\
ICL3    & $(24,10)$ & $(17,5)$   & $(15,5)$ & $(25,6)$ & $(13,6)$   & $(16,5)$ \\
\bottomrule
\end{tabular}
\end{subtable}
\begin{subtable}{\textwidth}
\vspace{1em}
\centering
\caption{ARIs for the estimated clustering}
\label{table_ari}
\begin{tabular}{ c | ccc | ccc | ccc}
\toprule
& \multicolumn{3}{c|}{$m=30$} & \multicolumn{3}{c|}{$m=50$} & \multicolumn{3}{c}{Alternative methods} \\
\midrule
 & $\hat{\mvec X}$ & $\tilde{\mvec X}$ & $\hat{\bm\Theta}$  & $\hat{\mvec X}$ & $\tilde{\mvec X}$ & $\hat{\bm\Theta}$ & Louvain & Paris & HLouvain \\
\midrule
ICL1    & $0.259$ & $0.324$ & $0.418$ & $0.262$ & $0.317$ & $0.418$ & $0.107$ & $0.082$ & $0.152$ \\
ICL2    & $0.441$ & $0.736$ & $0.938$  & $0.359$ & $0.743$ & $0.938$ & $0.488$ & $0.560$ & $0.602$ \\
ICL3 & $0.246$ & $0.342$ & $0.409$ & $0.269$ & $0.265$ & $0.364$ & $0.079$ & $0.032$ & $0.157$  \\
\bottomrule
\end{tabular}
\end{subtable}
\caption{Estimates of $(d,K)$ and ARIs for the embeddings $\hat{\mvec X}$, $\tilde{\mvec X}$ and $\hat{\bm\Theta}$ for $m\in\{30,50\}$ and alternative methodologies.}
\end{table}
\endgroup

Comparing Table~\ref{table_bic} for the two different values of $m$, it seems that 
the estimates of $K$ based on $\hat{\bm\Theta}$ 
are closer to the underlying true number of communities. 
In particular,
$K=3$ for ICL1, $K=4$ for ICL2, and $K=5$ 
for ICL3, based on the number of 
departments. 

Based on the estimates in Table~\ref{table_bic}, the estimated community allocations $z_i$ were obtained using \eqref{cluster_estimate}, and the adjusted Rand index was calculated using the department  
as labels. For further comparisons, the results were compared to other popular community detection methods: the Louvain algorithm \citep{Blondel08} adjusted for bipartite graphs \citep{Dugue15}, and the hierarchical Louvain (HLouvain) and Paris methods \citep{Bonald18}, all in their default implementation for bipartite graphs in the \textit{python} library \textit{scikit-network}. The results are reported in Table~\ref{table_ari}. Clearly, clustering on the embedding $\hat{\bm\Theta}$ outperforms the alternatives, including $\hat{\mvec X}$ and $\tilde{\mvec X}$, in all the three networks. In some cases, the improvement is substantial, for example in ICL2, where the proposed method reaches the score $0.938$, corresponding to only $9$ misclassified nodes out of $439$, particularly remarkable considering the lack of separation of the clusters in Figure~\ref{icl_scatterplots}. 

The results were confirmed by binomial paired sign tests on the difference between ARI scores for $N=25$ iterations of the community detection algorithms, which returned $p$-values $<10^{-4}$ in favour of the spherical coordinates estimator over the alternative methodologies. 

In terms of network structure, the results could be interpreted as follows: the computers connect to a set of shared college-wide web servers (for example, the virtual learning environment and the library services), and to services specific to the discipline carried out in each department. Furthermore, each machine has different levels of activity, which leads to heterogeneous within-community degree distributions. The SBM, corresponding to $\hat{\mvec X}$, only clusters the nodes based on their degree, whereas the DCSBM is able to uncover the departmental structure, in particular when the spherical coordinates estimator $\hat{\bm\Theta}$ is used.
Under the assumption that the networks were generated under a DCSBM, the results confirm that the estimate based on spherical coordinates proposed in this work appears to produce superior results when compared to the standard or row-normalised ASE.

\section{Conclusion}

In this article, a novel method for spectral clustering under the degree-corrected stochastic blockmodel has been proposed. The model is based on a transformation to spherical coordinates of the commonly used adjacency spectral embedding. Such a transformation seems more suited to Gaussian mixture modelling than the row-normalised embedding, which is widely used in the literature for spectral clustering. The proposed methodology is then incorporated within a simultaneous model selection scheme that allows the model dimension $d$ and the number of communities $K$ to be determined. The optimal values of $d$ and $K$ are chosen using the popular Bayesian information criterion. The framework also extends simply to include directed and bipartite graphs. Results on synthetic data and real-world computer networks show superior performance over competing methods.

\bibliographystyle{rss}
{\small
\bibliography{biblio}

\begin{thebibliography}{38}
\expandafter\ifx\csname natexlab\endcsname\relax\def\natexlab#1{#1}\fi
\expandafter\ifx\csname url\endcsname\relax
  \def\url#1{\texttt{#1}}\fi
\expandafter\ifx\csname urlprefix\endcsname\relax\def\urlprefix{URL: }\fi

\bibitem[{Alanis-Lobato et~al.(2016)Alanis-Lobato, Mier and
  Andrade-Navarro}]{Lobato16}
Alanis-Lobato, G., Mier, P. and Andrade-Navarro, M.~A. (2016) Efficient
  embedding of complex networks to hyperbolic space via their {Laplacian}.
\newblock \textit{Scientific Reports}, \textbf{6}, 30108.

\bibitem[{Amini et~al.(2013)Amini, Chen, Bickel and Levina}]{Amini13}
Amini, A.~A., Chen, A., Bickel, P.~J. and Levina, E. (2013) Pseudo-likelihood
  methods for community detection in large sparse networks.
\newblock \textit{Annals of Statistics}, \textbf{41}, 2097--2122.

\bibitem[{Athreya et~al.(2016)Athreya, Priebe, Tang, Lyzinski, Marchette and
  Sussman}]{Athreya16}
Athreya, A., Priebe, C.~E., Tang, M., Lyzinski, V., Marchette, D.~J. and
  Sussman, D.~L. (2016) A limit theorem for scaled eigenvectors of random dot
  product graphs.
\newblock \textit{Sankhya A}, \textbf{78}, 1--18.

\bibitem[{Blondel et~al.(2008)Blondel, Guillaume, Lambiotte and
  Lefebvre}]{Blondel08}
Blondel, V.~D., Guillaume, J.-L., Lambiotte, R. and Lefebvre, E. (2008) Fast
  unfolding of communities in large networks.
\newblock \textit{Journal of Statistical Mechanics: Theory and Experiment},
  P10008.

\bibitem[{Bonald et~al.(2018)Bonald, Charpentier, Galland and
  Hollocou}]{Bonald18}
Bonald, T., Charpentier, B., Galland, A. and Hollocou, A. (2018) Hierarchical
  graph clustering based on node pair sampling.
\newblock In \textit{Proceedings of the 14th International Workshop on Mining
  and Learning with Graphs (MLG)}.

\bibitem[{Braun and Bonfrer(2011)}]{Braun11}
Braun, M. and Bonfrer, A. (2011) Scalable inference of customer similarities
  from interactions data using {Dirichlet} processes.
\newblock \textit{Marketing science}, \textbf{30}, 513--531.

\bibitem[{Chaudhuri et~al.(2012)Chaudhuri, Chung and Tsiatas}]{Chaudhuri12}
Chaudhuri, K., Chung, F. and Tsiatas, A. (2012) Spectral clustering of graphs
  with general degrees in the extended planted partition model.
\newblock In \textit{Proceedings of the 25th Annual Conference on Learning
  Theory}, vol.~23.

\bibitem[{Chen et~al.(2018)Chen, Li and Xu}]{Chen18}
Chen, Y., Li, X. and Xu, J. (2018) Convexified modularity maximization for
  degree-corrected stochastic block models.
\newblock \textit{Annals of Statistics}, \textbf{46}, 1573--1602.

\bibitem[{Dempster et~al.(1977)Dempster, Laird and Rubin}]{Dempster77}
Dempster, A., Laird, N. and Rubin, D. (1977) Maximum likelihood from incomplete
  data via the {EM} algorithm.
\newblock \textit{Journal of the Royal Statistical Society, Series B},
  \textbf{39}, 1--38.

\bibitem[{Dugu{\'e} and Perez(2015)}]{Dugue15}
Dugu{\'e}, N. and Perez, A. (2015) Directed {Louvain}: maximizing modularity in
  directed networks.
\newblock \textit{Tech. Rep. hal-01231784}, Universit{\'e} d'Orl{\'e}ans.

\bibitem[{Fraley and Raftery(2002)}]{Raftery02}
Fraley, C. and Raftery, A.~E. (2002) Model-based clustering, discriminant
  analysis, and density estimation.
\newblock \textit{Journal of the American Statistical Association},
  \textbf{97}, 611--631.

\bibitem[{Gao et~al.(2018)Gao, Ma, Zhang and Zhou}]{Gao18}
Gao, C., Ma, Z., Zhang, A.~Y. and Zhou, H.~H. (2018) Community detection in
  degree-corrected block models.
\newblock \textit{Annals of Statistics}, \textbf{46}, 2153--2185.

\bibitem[{Gulikers et~al.(2017)Gulikers, Lelarge and
  Massouli{\'e}}]{Gulikers17}
Gulikers, L., Lelarge, M. and Massouli{\'e}, L. (2017) A spectral method for
  community detection in moderately sparse degree-corrected stochastic block
  models.
\newblock \textit{Advances in Applied Probability}, \textbf{49}, 686--721.

\bibitem[{Holland et~al.(1983)Holland, Laskey and Leinhardt}]{Holland83}
Holland, P.~W., Laskey, K.~B. and Leinhardt, S. (1983) Stochastic blockmodels:
  First steps.
\newblock \textit{Social Networks}, \textbf{5}, 109--137.

\bibitem[{Hubert and Arabie(1985)}]{Hubert85}
Hubert, L. and Arabie, P. (1985) Comparing partitions.
\newblock \textit{Journal of Classification}, \textbf{2}, 193--218.

\bibitem[{Jin(2015)}]{Jin15}
Jin, J. (2015) Fast community detection by {SCORE}.
\newblock \textit{Annals of Statistics}, \textbf{43}, 57--89.

\bibitem[{{Jones} and {Rubin-Delanchy}(2020)}]{Jones20}
{Jones}, A. and {Rubin-Delanchy}, P. (2020) {The multilayer random dot product
  graph}.
\newblock \textit{arXiv e-prints}.

\bibitem[{Karrer and Newman(2011)}]{Karrer11}
Karrer, B. and Newman, M. E.~J. (2011) Stochastic blockmodels and community
  structure in networks.
\newblock \textit{Physical Review E}, \textbf{83}.

\bibitem[{Krioukov et~al.(2010)Krioukov, Papadopoulos, Kitsak, Vahdat and
  Bogu\~n\'a}]{Krioukov10}
Krioukov, D., Papadopoulos, F., Kitsak, M., Vahdat, A. and Bogu\~n\'a, M.
  (2010) Hyperbolic geometry of complex networks.
\newblock \textit{Physical Review E}, \textbf{82}.

\bibitem[{Lei and Rinaldo(2015)}]{Lei15}
Lei, J. and Rinaldo, A. (2015) Consistency of spectral clustering in stochastic
  block models.
\newblock \textit{Annals of Statistics}, \textbf{43}, 215--237.

\bibitem[{von Luxburg(2007)}]{vonLuxburg07}
von Luxburg, U. (2007) A tutorial on spectral clustering.
\newblock \textit{Statistics and Computing}, \textbf{17}.

\bibitem[{Mardia(1970)}]{Mardia70}
Mardia, K.~V. (1970) Measures of multivariate skewness and kurtosis with
  applications.
\newblock \textit{Biometrika}, \textbf{57}, 519--530.

\bibitem[{McCormick and Zheng(2015)}]{McCormick15}
McCormick, T.~H. and Zheng, T. (2015) Latent surface models for networks using
  aggregated relational data.
\newblock \textit{Journal of the American Statistical Association},
  \textbf{110}, 1684--1695.

\bibitem[{Meyer(2000)}]{Meyer00}
Meyer, C.~D. (2000) \textit{Matrix analysis and applied linear algebra}.
\newblock SIAM.

\bibitem[{Neil et~al.(2013)Neil, Hash, Brugh, Fisk and Storlie}]{Neil13}
Neil, J., Hash, C., Brugh, A., Fisk, M. and Storlie, C.~B. (2013) Scan
  statistics for the online detection of locally anomalous subgraphs.
\newblock \textit{Technometrics}, \textbf{55}, 403--414.

\bibitem[{Ng et~al.(2001)Ng, Jordan and Weiss}]{Ng01}
Ng, A.~Y., Jordan, M.~I. and Weiss, Y. (2001) On spectral clustering: Analysis
  and an algorithm.
\newblock In \textit{Proceedings of the 14th International Conference on Neural
  Information Processing Systems: Natural and Synthetic}, 849--856.

\bibitem[{Peng and Carvalho(2016)}]{Peng16}
Peng, L. and Carvalho, L. (2016) Bayesian degree-corrected stochastic
  blockmodels for community detection.
\newblock \textit{Electronic Journal of Statistics}, \textbf{10}, 2746--2779.

\bibitem[{Qin and Rohe(2013)}]{Qin13}
Qin, T. and Rohe, K. (2013) Regularized spectral clustering under the
  degree-corrected stochastic blockmodel.
\newblock In \textit{Proceedings of the 26th International Conference on Neural
  Information Processing Systems}, vol.~2, 3120--3128.

\bibitem[{Raftery and Dean(2006)}]{Dean06}
Raftery, A.~E. and Dean, N. (2006) Variable selection for model-based
  clustering.
\newblock \textit{Journal of the American Statistical Association},
  \textbf{101}, 168--178.

\bibitem[{Rohe et~al.(2016)Rohe, Qin and Yu}]{Rohe16}
Rohe, K., Qin, T. and Yu, B. (2016) Co-clustering directed graphs to discover
  asymmetries and directional communities.
\newblock \textit{Proceedings of the National Academy of Sciences}.

\bibitem[{{Rubin-Delanchy} et~al.(2017){Rubin-Delanchy}, {Cape}, {Tang} and
  {Priebe}}]{RubinDelanchy17}
{Rubin-Delanchy}, P., {Cape}, J., {Tang}, M. and {Priebe}, C.~E. (2017) {A
  statistical interpretation of spectral embedding: the generalised random dot
  product graph}.
\newblock \textit{arXiv e-prints}.

\bibitem[{{Sanna Passino} and Heard(2020)}]{SannaPassino19}
{Sanna Passino}, F. and Heard, N.~A. (2020) Bayesian estimation of the latent
  dimension and communities in stochastic blockmodels.
\newblock \textit{Statistics and Computing}, \textbf{30}, 1291--1307.

\bibitem[{{Sussman} et~al.(2014){Sussman}, {Tang} and {Priebe}}]{Sussman14}
{Sussman}, D.~L., {Tang}, M. and {Priebe}, C.~E. (2014) Consistent latent
  position estimation and vertex classification for random dot product graphs.
\newblock \textit{IEEE Transactions on Pattern Analysis and Machine
  Intelligence}, \textbf{36}, 48--57.

\bibitem[{Tang and Priebe(2018)}]{Tang18}
Tang, M. and Priebe, C.~E. (2018) Limit theorems for eigenvectors of the
  normalized {Laplacian} for random graphs.
\newblock \textit{Annals of Statistics}, \textbf{46}, 2360--2415.

\bibitem[{{Yang} et~al.(2020){Yang}, {Priebe}, {Park} and {Marchette}}]{Yang19}
{Yang}, C., {Priebe}, C.~E., {Park}, Y. and {Marchette}, D.~J. (2020)
  Simultaneous dimensionality and complexity model selection for spectral graph
  clustering.
\newblock \textit{Journal of Computational and Graphical Statistics},
  \textbf{30}, 422--441.

\bibitem[{Young and Scheinerman(2007)}]{Young07}
Young, S.~J. and Scheinerman, E.~R. (2007) Random dot product graph models for
  social networks.
\newblock In \textit{Algorithms and Models for the Web-Graph}, 138--149.

\bibitem[{Zhao et~al.(2012)Zhao, Levina and Zhu}]{Zhao12}
Zhao, Y., Levina, E. and Zhu, J. (2012) Consistency of community detection in
  networks under degree-corrected stochastic block models.
\newblock \textit{Annals of Statistics}, \textbf{40}, 2266--2292.

\bibitem[{Zhu and Ghodsi(2006)}]{Zhu06}
Zhu, M. and Ghodsi, A. (2006) Automatic dimensionality selection from the scree
  plot via the use of profile likelihood.
\newblock \textit{Computational Statistics \& Data Analysis}, \textbf{51}, 918
  -- 930.

\end{thebibliography}
}

\appendix

\section{A central limit theorem on spherical coordinates} \label{asymptotic}

In this section, it is demonstrated that, for $d=2$, the spherical coordinates obtained from ASE asymptotically converge to the spherical coordinates of the true underlying latent positions, with Gaussian error. This property is particularly important, since points belonging to the same community share the \textit{same spherical coordinates} under the DCSBM.

Consider the covariance $\bm\Sigma_k(\rho)$ appearing in the ASE-CLT \eqref{ase-clt}, which holds under the assumption of $d$ fixed and known.
The covariance matrix $\bm\Sigma_k(\rho)$ can be calculated explicitly, and arises as a corollary of the ASE-CLT, taking the RDPG inner product distribution to be the product measure
$
F = G_\rho \otimes \sum_{k=1}^K \psi_k \delta_{\bm\mu_k},\ \sum_{k=1}^K \psi_k=1,\ \psi_k\geq 0,
$
where $G_\rho$ is the distribution of the degree-correction parameters, and $\delta_\cdot$ is the Dirac's delta measure. Letting $\bm\xi\sim F$ be a $d$-dimensional random vector, the $k$-th community covariance matrix then takes the form
$
\bm\Sigma_k(\rho) = \bm\Delta^{-1}\mathbb E\{\rho\bm\mu_k^\intercal\bm\xi(1-\rho\bm\mu_k^\intercal\bm\xi)\bm\xi\bm\xi^\intercal\}\bm\Delta^{-1},
$
where $\bm\Delta=\mathbb E(\bm\xi\bm\xi^\intercal)\in\mathbb R^{d\times d}$ is the second moment matrix, assumed to be invertible. 

The ASE-CLT \eqref{ase-clt} provides the theoretical framework for establishing a central limit theorem for the spherical coordinates estimator under the DCSBM for $d=2$. Let $\hat{\theta}=f_2(\vec x)$ be the spherical coordinates for the true latent position $\vec x\in\mathbb R^2$, and $\hat{\theta}^{(n)}=f_2(\hat{\vec x}^{(n)})$ be the corresponding estimator calculated from $\hat{\vec x}^{(n)}\in\mathbb R^2$. Note that for $d=2$, $\mvec Q^{(n)}\in\mathbb O(2)$, the orthogonal group in two dimensions, which consists in rotation and reflection matrices:
\begin{align}
\mvec Q_{\mathrm{rot}}(\varphi) = 
\begin{bmatrix} \cos(\varphi) & \sin(\varphi) \\ -\sin(\varphi) & \cos(\varphi) \end{bmatrix}, &&
\mvec Q_{\mathrm{ref}}(\varphi) = 
\begin{bmatrix} -\cos(\varphi) & \sin(\varphi) \\ \sin(\varphi) & \cos(\varphi) \end{bmatrix}.
\end{align}
The rotation matrix $\mvec Q_{\mathrm{rot}}(\varphi)$ applied to $\vec x\in\mathbb R^2$ rotates the vector by an angle $\varphi$ (taken clockwise) with respect to the second axis, whereas a reflection matrix $\mvec Q_{\mathrm{ref}}(\varphi)$ reflects the vector with respect to a line passing through the origin, with angle $\varphi/2$ (taken clockwise) to the second axis. 
It follows that any matrix $\mvec Q^{(n)}\in\mathbb O(2)$ appearing in the ASE-CLT \eqref{ase-clt} is uniquely mapped to an angle $\varphi^{(n)}\in[-2\pi,2\pi)$ such that:
\begin{equation}
f_2(\mvec Q^{(n)}\hat{\vec x}^{(n)})= (f_2(\hat{\vec x}^{(n)}) + \varphi^{(n)}) \bmod 2\pi = (\hat\theta^{(n)}  + \varphi^{(n)}) \bmod 2\pi.
\end{equation}
The ASE-CLT \eqref{ase-clt} can be extended to spherical coordinates by the multivariate delta method, with transformation function $f_2(\cdot)$, see \eqref{cart_to_sphere}.
The multivariate delta method 
establishes that, conditional on the community allocation, correction parameter and orthogonal matrices $\mvec Q^{(n)}\in\mathbb O(2)$, the spherical coordinates are asymptotically Gaussian:
\begin{equation}
\lim_{n\to\infty} \mathbb P\left\{\sqrt{n}\left(\hat\theta^{(n)} - \theta - \varphi^{(n)}\right)\leq v\ \Big\vert\ z = k, \rho\right\} \to \Phi_1\left\{v,\bm\nabla_{\vec x}(\theta)^\intercal\bm\Sigma_k(\rho)\bm\nabla_{\vec x}(\theta)\right\},
\end{equation}
where $v\in\mathbb R$, and $\bm\nabla_{\vec x}(\theta)$ is the gradient of $\theta$ with respect to $\vec x$, under the assumption that $(\hat\theta^{(n)}  + \varphi^{(n)}) \bmod 2\pi = \hat\theta^{(n)}  + \varphi^{(n)}$.
The gradient is:
\begin{equation}
\bm\nabla_{\vec x}(\theta) = \left(\der{\theta}{x_1}, \der{\theta}{x_2}\right)^\intercal = \left(0,-2\cdot\mathrm{sign}(x_1)\sqrt{1- x_2^2/\|\vec x\|^2}\big/\|\vec x\|\right)^\intercal. \label{gradient}
\end{equation}

\section{Empirical model validation: additional simulations} \label{app_sim}

\subsection{Asymptotic behaviour}

The proposed model is further validated by a study with increasing values for the number of nodes $n$. $N=\numprint{1000}$ graphs are simulated using the same configuration as the simulation in Figure~\ref{simulation_results}, for $n\in\{100,200,500,1000,2000\}$. Figure~\ref{simulation_n} reports the boxplots for the different values of $n$ for three of the estimated parameters, suggesting that the asymptotic behaviour of the embedding is consistent with the modelling choices in \eqref{full_model}. In particular, three assumptions are checked: means centred at $\pi$ in the last $m-d$ components (\textit{cf.} Figure~\ref{means_n}), independence between the initial $d$ and last $m-d$ components (\textit{cf.} Figure~\ref{covs_n}), and independence within the last $m-d$ components (\textit{cf.} Figure~\ref{covs_n2}). 

\begingroup

\begin{figure}[!t]
\centering
\begin{subfigure}[t]{.325\textwidth}
\centering
\caption{Boxplots of $\hat\vartheta_{k,2}$}
\includegraphics[width=.975\textwidth]{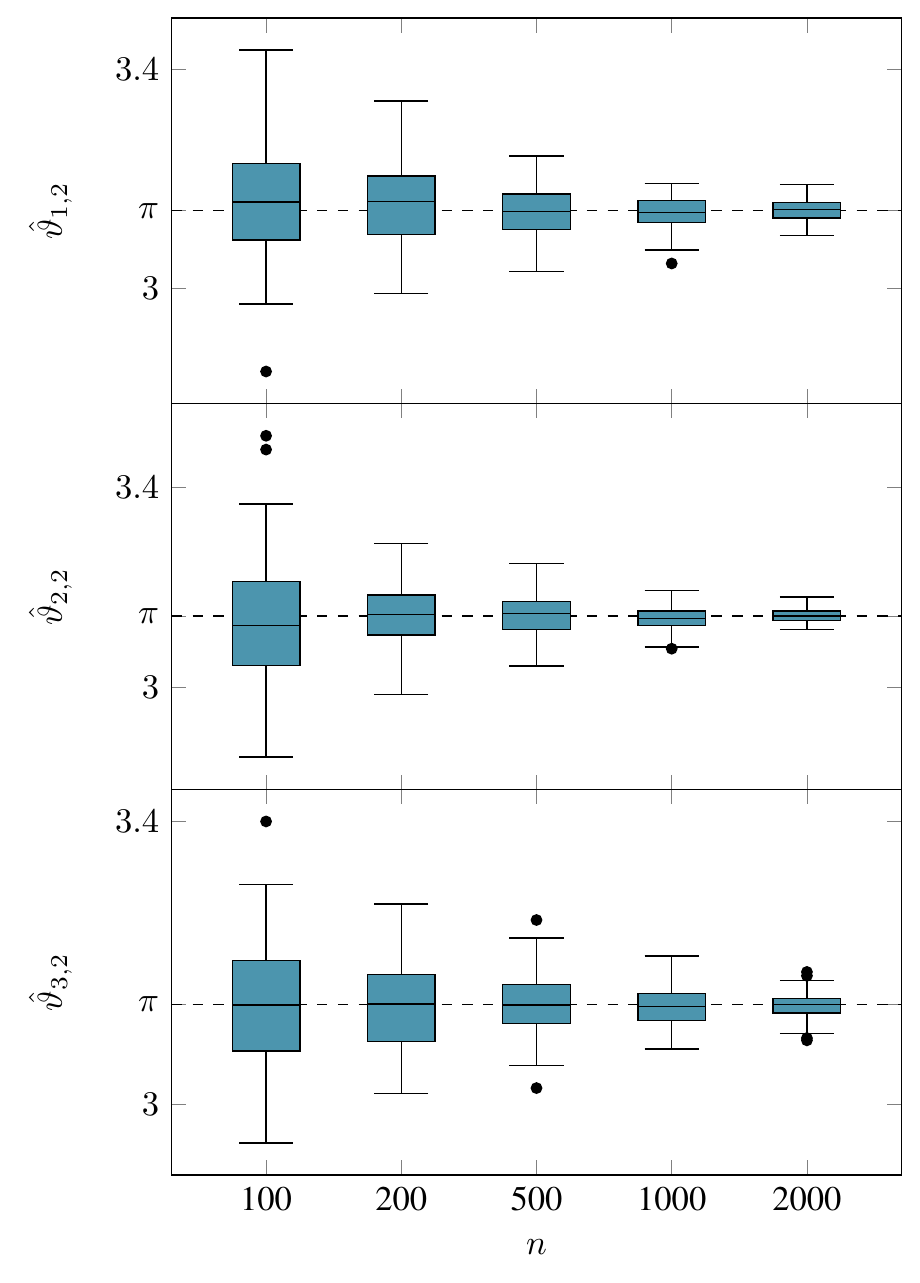}
\label{means_n}
\end{subfigure}
\begin{subfigure}[t]{.325\textwidth}
\centering
\caption{Boxplots of $[\hat{\bm\Sigma}_k]_{1,3}$}
\includegraphics[width=.975\textwidth]{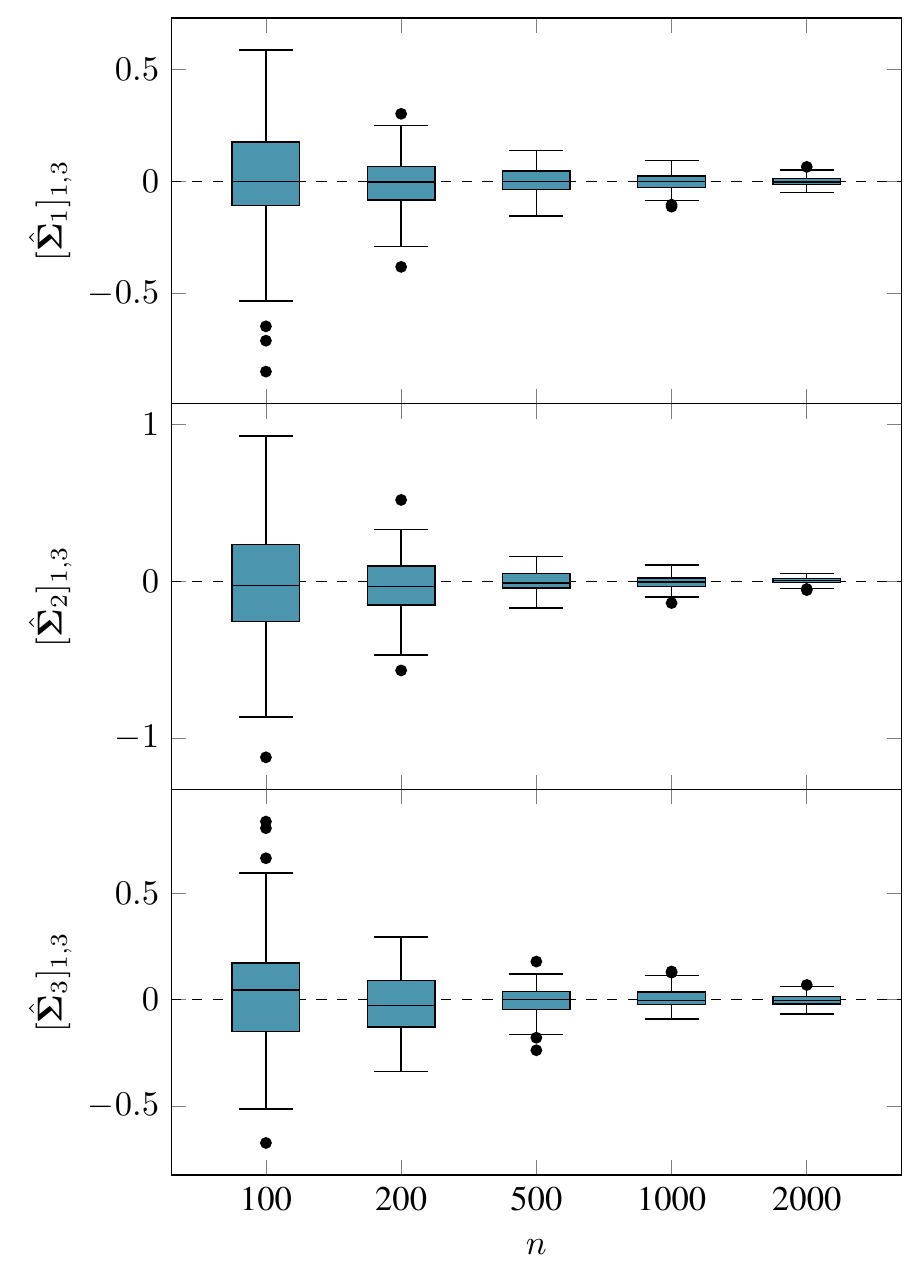}
\label{covs_n}
\end{subfigure}
\begin{subfigure}[t]{.325\textwidth}
\centering
\caption{Boxplots of $[\hat{\bm\Sigma}_k]_{3,4}$}
\includegraphics[width=.975\textwidth]{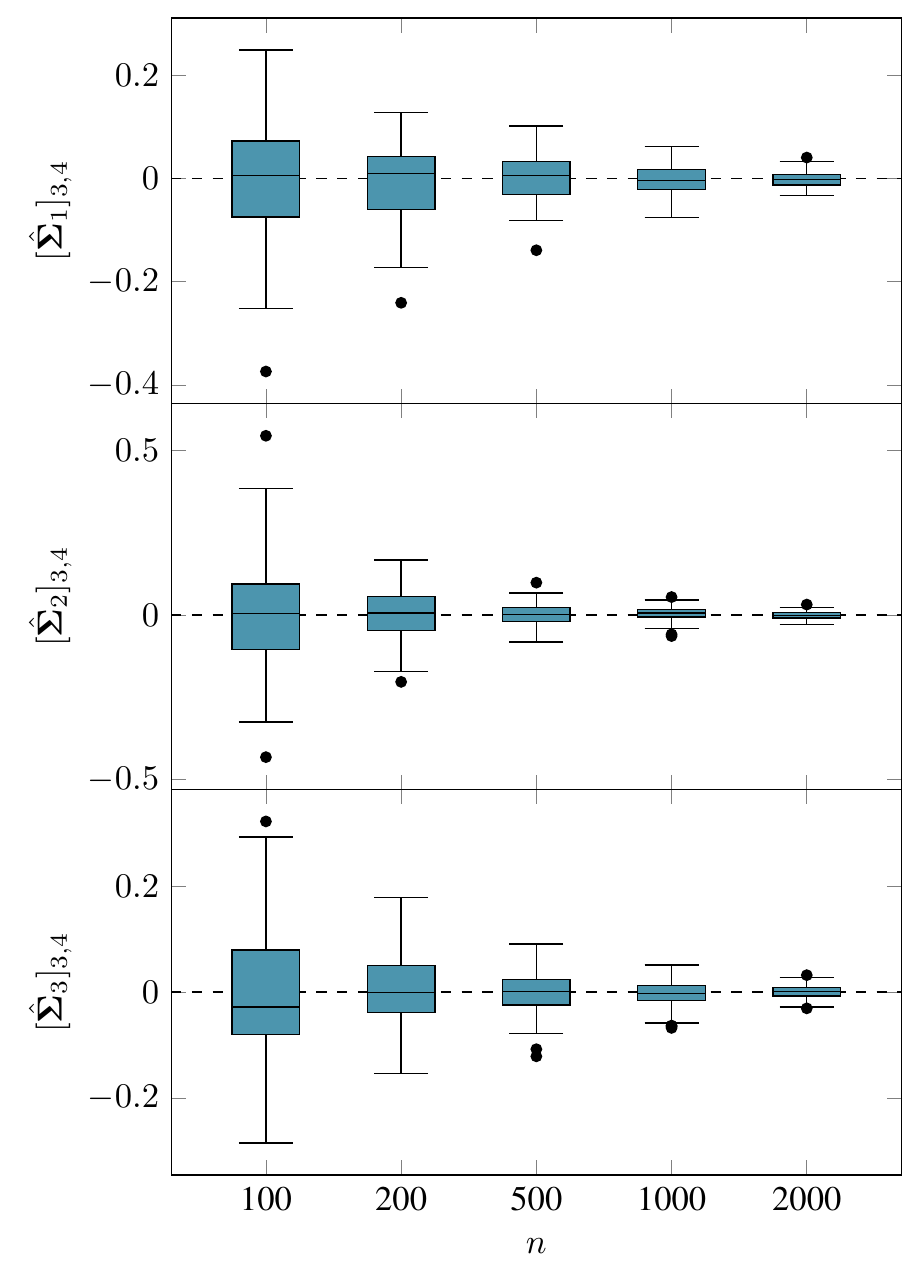}
\label{covs_n2}
\end{subfigure}
\caption{Boxplots for $N=\numprint{1000}$ simulations of a degree-corrected stochastic blockmodel with varying number of nodes $n$, $K=3$, equal number of nodes allocated to each group, and $\mvec B$ described in \eqref{blockmatrix}, corrected by parameters $\rho_i$ sampled from a $\mathrm{Uniform}(0,1)$ distribution.}
\label{simulation_n}
\end{figure}
\endgroup

\subsection{Changes in the correlation between blocks}

The same model assumptions are also checked with different values of the correlation between blocks. $N=\numprint{100}$ graphs with $K=2$ equal-sized communities and $n=\numprint{500}$ nodes are simulated from a DCSBM with $B_{11}=0.5$, $B_{22}=0.35$ and $B_{12}=B_{21}=r$, $r\in\{0.05,0.1,0.15,0.2,0.25,0.3,0.35\}$, corrected by $\rho_i\sim\mathrm{Uniform}(0,1)$. The results plotted in Figure~\ref{simulation_rho} suggest that the assumptions in model \eqref{full_model} are robust to changes in the correlation between the blocks.
 
\begingroup

\begin{figure}[!h]
\centering
\begin{subfigure}[t]{.325\textwidth}
\centering
\caption{Boxplots of $\hat\vartheta_{k,2}$}
\includegraphics[width=.975\textwidth]{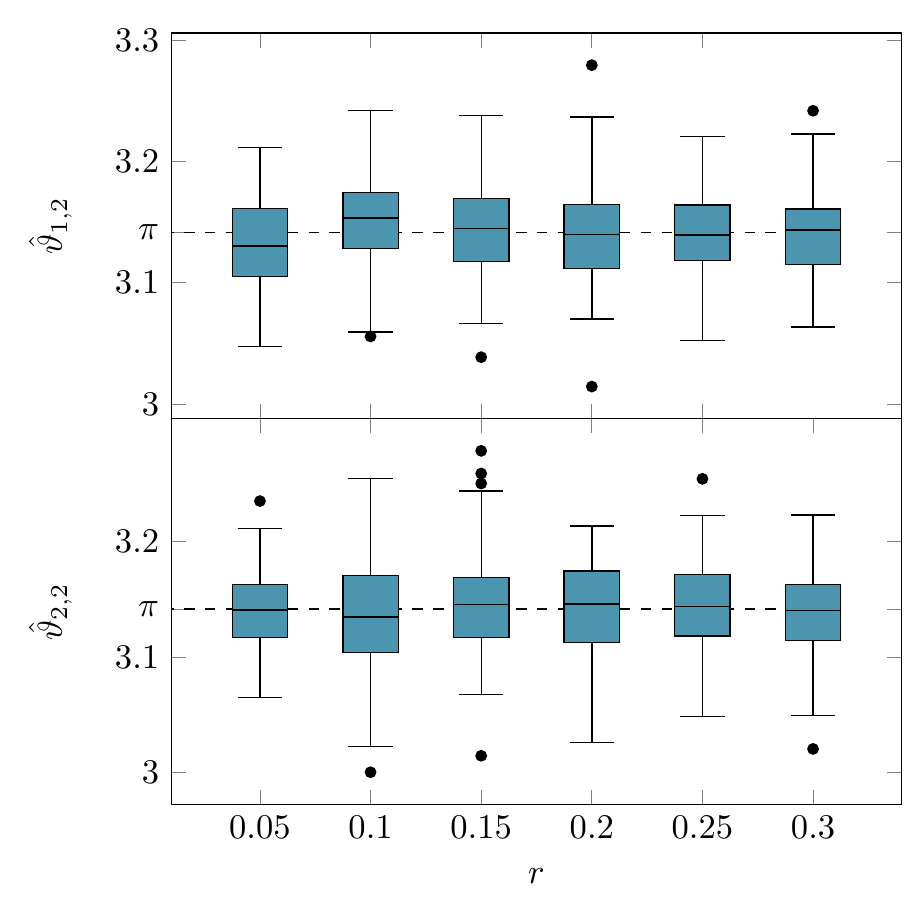}
\label{means_rho}
\end{subfigure}
\begin{subfigure}[t]{.325\textwidth}
\centering
\caption{Boxplots of $[\hat{\bm\Sigma}_k]_{1,3}$}
\includegraphics[width=.975\textwidth]{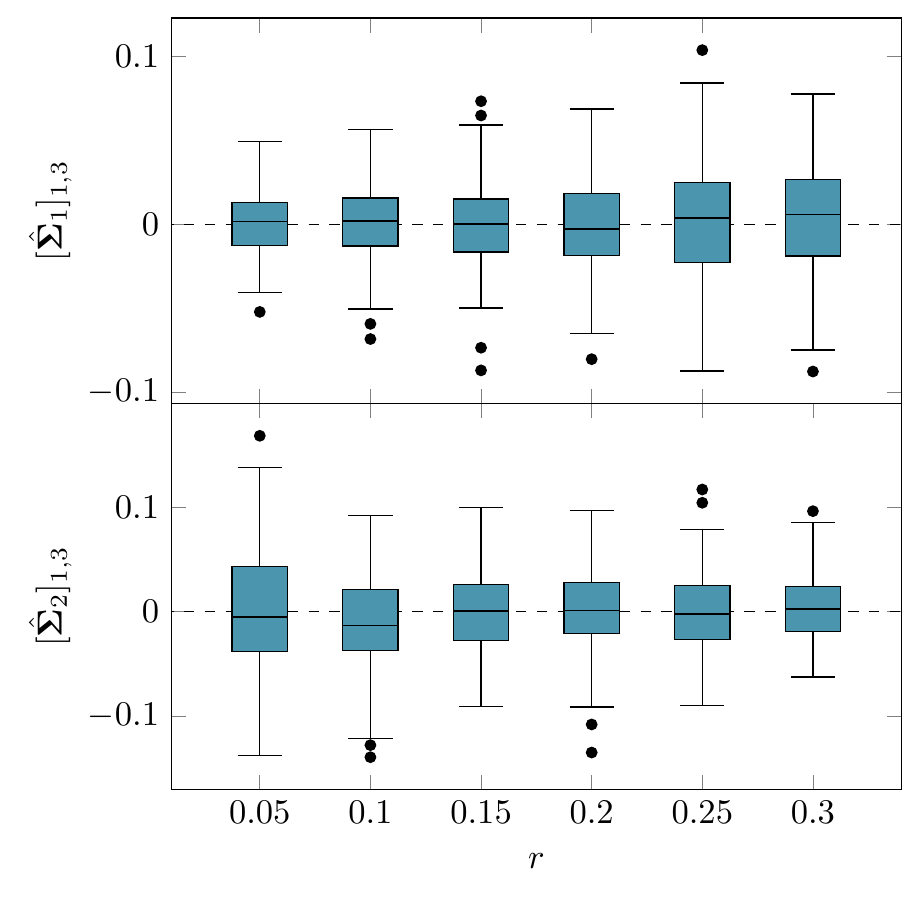}
\label{covs_rho}
\end{subfigure}
\begin{subfigure}[t]{.325\textwidth}
\centering
\caption{Boxplots of $[\hat{\bm\Sigma}_k]_{3,4}$}
\includegraphics[width=.975\textwidth]{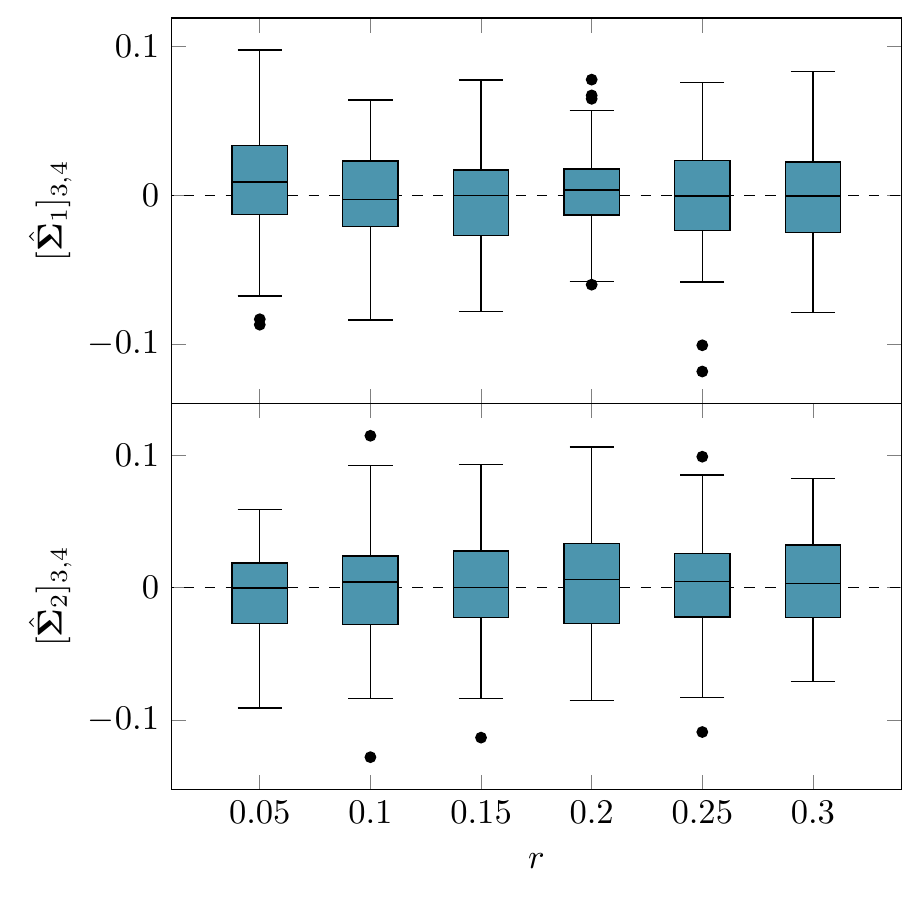}
\label{covs_rho2}
\end{subfigure}
\caption{Boxplots for $N=\numprint{100}$ simulations of a degree-corrected stochastic blockmodel with $n=500$, $K=2$, equal number of nodes allocated to each group, $B_{11}=0.5$, $B_{22}=0.35$ and $B_{12}=B_{21}=r$, for different values of $r$, corrected by parameters $\rho_i$ sampled from a $\mathrm{Uniform}(0,1)$ distribution.}
\label{simulation_rho}
\end{figure}
\endgroup

\end{document}